\documentclass[10pt,twocolumn,letterpaper]{article}

\usepackage[pagenumbers]{cvpr} 

\usepackage{multirow}
\usepackage{tabularx}
\usepackage{algorithm}
\usepackage{algorithmic}
\usepackage{listings}
\usepackage[most]{tcolorbox}

\newtcblisting{promptbox}[1]{
    listing only,
    breakable,
    colback=black!2,
    colframe=black!20,
    colbacktitle=black!4,
    coltitle=black,
    fonttitle=\bfseries\small,
    title={#1},
    boxrule=0.4pt,
    arc=1pt,
    left=6pt,
    right=6pt,
    top=5pt,
    bottom=5pt,
    before skip=6pt,
    after skip=8pt,
    listing options={
        basicstyle=\footnotesize\ttfamily,
        breaklines=true,
        breakatwhitespace=true,
        breakautoindent=false,
        breakindent=0pt,
        columns=fullflexible,
        keepspaces=true,
        showstringspaces=false
    }
}

\usepackage[table]{xcolor}
\definecolor{genevalblue}{RGB}{232,239,250}
\definecolor{genevalgreen}{RGB}{232,245,225}
\definecolor{genevalpink}{RGB}{252,232,238}
\definecolor{defaultrow}{RGB}{238,243,250}
\definecolor{cvprblue}{rgb}{0.21,0.49,0.74}
\usepackage[pagebackref,breaklinks,colorlinks,allcolors=cvprblue]{hyperref}

\def\confName{CVPR}
\def\confYear{2027}

\title{PromptSMC: Enabling Particle Exploration in Flow Matching \\ via LLM-Based Prompt Rejuvenation}

\author{Wenyi Lian\\
Uppsala University\\
{\tt\small wenyi.lian@it.uu.se}
\and
Zheng Zhao\\
Link\"{o}ping University\\
{\tt\small zheng.zhao@liu.se}
\and
Thomas B. Sch{\"o}n\\
Uppsala University\\
{\tt\small thomas.schon@uu.se }
}

\begin{document}
\maketitle
\begin{abstract}
  This paper is concerned with prompt-image alignment at inference time through exploration of generative trajectories and dynamical prompt rejuvenation. 
  We leverage sequential Monte Carlo (SMC), to offer a gradient-free and principled framework for this purpose through reward-guided trajectory selection. 
  More specifically, we introduce LLM-based prompt rejuvenation, allowing for efficient particle exploration in SMC samplers along the diffusion time in flow matching models. 
  The proposed method, \textbf{PromptSMC}, retains the selected latent states and uses a pretrained LLM to dynamically revise the prompts of duplicated particles under semantic-preservation constraints. These revised prompts induce alternative conditional velocity fields, allowing particles to explore different continuations from a shared intermediate state. The method further uses a reward model to evaluate each particle's alignment with the original prompt, and the resulting scores guide particles through resampling. PromptSMC thus integrates prompt-based exploration with reward-guided sampling, achieving faithful prompt-image alignment without additional noise injection, model training, or reward gradients. Our experiments and ablation studies across multiple benchmarks and backbones demonstrate the effectiveness of the method for text-to-image flow matching. Code is available at \url{https://github.com/shermanlian/PromptSMC}.
\end{abstract}

\section{Introduction}
\label{sec:introduction}

Flow matching (FM) has become an effective framework for text-to-image generation, producing images by integrating a learned conditional velocity field with an ordinary differential equation (ODE) solver~\cite{lipman2022flow,liu2022flow,esser2024scaling}. 
While contemporary flow models produce visually compelling images, accurately capturing object counts, attributes, and spatial relationships described in the input text remains challenging~\cite{ghosh2023geneval,huang2023t2i}. To address it, inference-time scaling improves the alignment by allocating additional computation to the sampling process~\cite{xie2025sana,ma2025inference,jain2026diffusion}. Particle-based approaches, such as sequential Monte Carlo (SMC)~\cite{doucet2001introduction,naesseth2019elements}, provide a statistically principled framework for this purpose by maintaining multiple generation processes, referred to as \emph{particles}, and selecting promising trajectories according to intermediate reward signals, enabling steering of the generation without necessarily requiring reward gradients~\cite{wu2023practical,singhal2025general}.

However, applying particle-based approaches to flow matching models presents an exploration challenge: under deterministic ODE sampling, particles replicated from the same state follow identical subsequent trajectories when their textual conditions remain unchanged. Consequently, allocating additional particles to a promising state does not produce alternative continuations, limiting exploration and potential performance gains. Existing approaches facilitate exploration by introducing stochasticity through SDE-based sampling~\cite{holderrieth2025glass,kim2026inference} and trajectory perturbation~\cite{stecklov2025inference,lian2026testtime}, or by pruning less promising trajectories~\cite{yu2025rethinking}. While these approaches focus on sampling dynamics or trajectory selection, we explore a complementary direction: \emph{can we enable particle exploration and improve prompt-image alignment by adapting the textual conditions while retaining deterministic ODE sampling?}

\begin{figure}
    \centering
    \includegraphics[width=1.\linewidth]{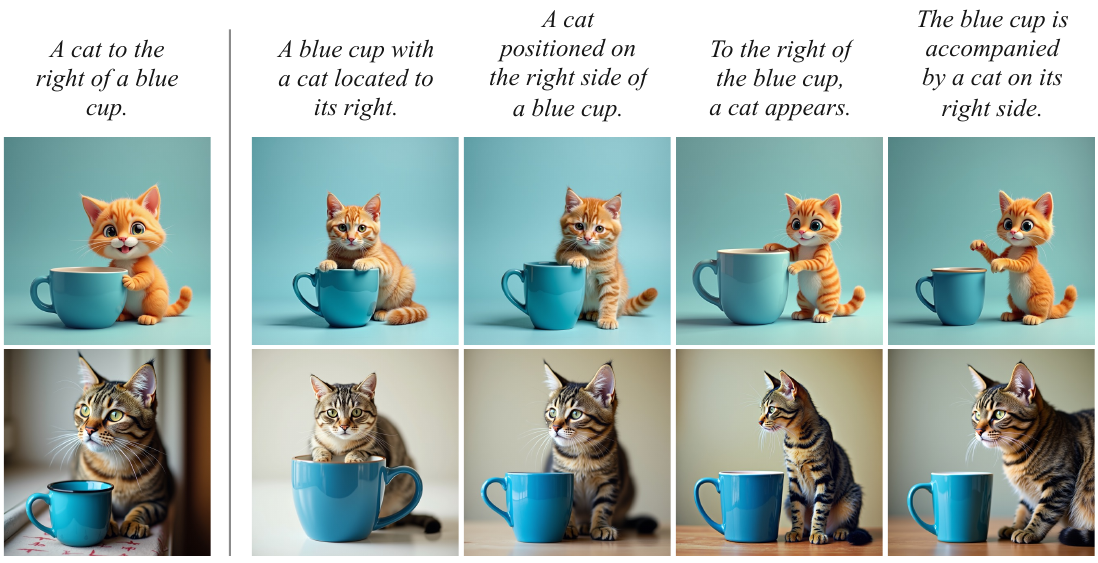}
    \caption{Prompt perturbation leads to different results. Each row represents images generated from the same latent noise. The same pretrained LLM~\cite{bai2025qwen3} is used for prompt rewriting.}
    \label{fig:latent-prompts}
    \vspace{-0.1in}
\end{figure}

Our key observation is that a text-to-image flow model predicts a \emph{text-conditioned} velocity field. Even at the same latent state, perturbing textual conditions can induce different velocities and subsequent trajectories. Prompt revision can thus enable particles to explore alternative continuations without perturbing their latent states. In particular, pretrained large language models (LLMs)~\cite{grattafiori2024llama,bai2025qwen3} can generate prompt variants under instructions to preserve the original meaning. These variants provide different generation conditions for expressing the same user request. \cref{fig:latent-prompts} illustrates such prompt variants and the corresponding images generated from the same initial noise.

In this paper, we introduce \textbf{PromptSMC}, which integrates LLM-based prompt rejuvenation with SMC to enable particle exploration in text-to-image flow matching.
More precisely, PromptSMC dynamically maintains multiple interacting particles in parallel. At each resampling step, a predefined reward model evaluates intermediate image estimates against the original user prompt, and the resulting scores determine the particle weights for resampling. The method then uses a pretrained LLM to generate prompt variants for duplicated particles. For each selected parent particle, one descendant particle retains its inherited prompt, while additional copies explore the revised conditions. These descendants continue from the shared latent state under different conditional velocity fields, enabling alternative trajectories to emerge after resampling. By combining prompt-based exploration with reward-guided selection, PromptSMC steers generation toward improved alignment while retaining the pretrained flow model and its deterministic ODE solver, without additional noise injection, model training, or reward gradients. \cref{fig:teaser} presents a qualitative comparison with FLUX.1-dev~\cite{labs2025flux}, highlighting the prompt-image alignment achieved by PromptSMC.

Our contributions are as follows.
\begin{itemize}
    \item We introduce LLM-based prompt rejuvenation, which enables particle exploration in flow matching by dynamically adapting textual conditions while retaining intermediate latent states and deterministic ODE propagation.
    \item We propose PromptSMC, combining prompt rejuvenation with reward-guided particle resampling for faithful prompt-image alignment at inference time without additional model training or reward gradients.
    \item We evaluate our method on three text-to-image flow matching models: SD3~\cite{esser2024scaling}, FLUX.1-dev~\cite{labs2025flux}, and Z-Image~\cite{cai2025z}. The experimental results on GenEval~\cite{ghosh2023geneval} and T2I-CompBench~\cite{huang2023t2i} demonstrate that PromptSMC consistently improves prompt-image alignment across different backbones and benchmarks. 
\end{itemize}
\begin{figure*}[t]
  \begin{center}
    \centerline{\includegraphics[width=\linewidth]{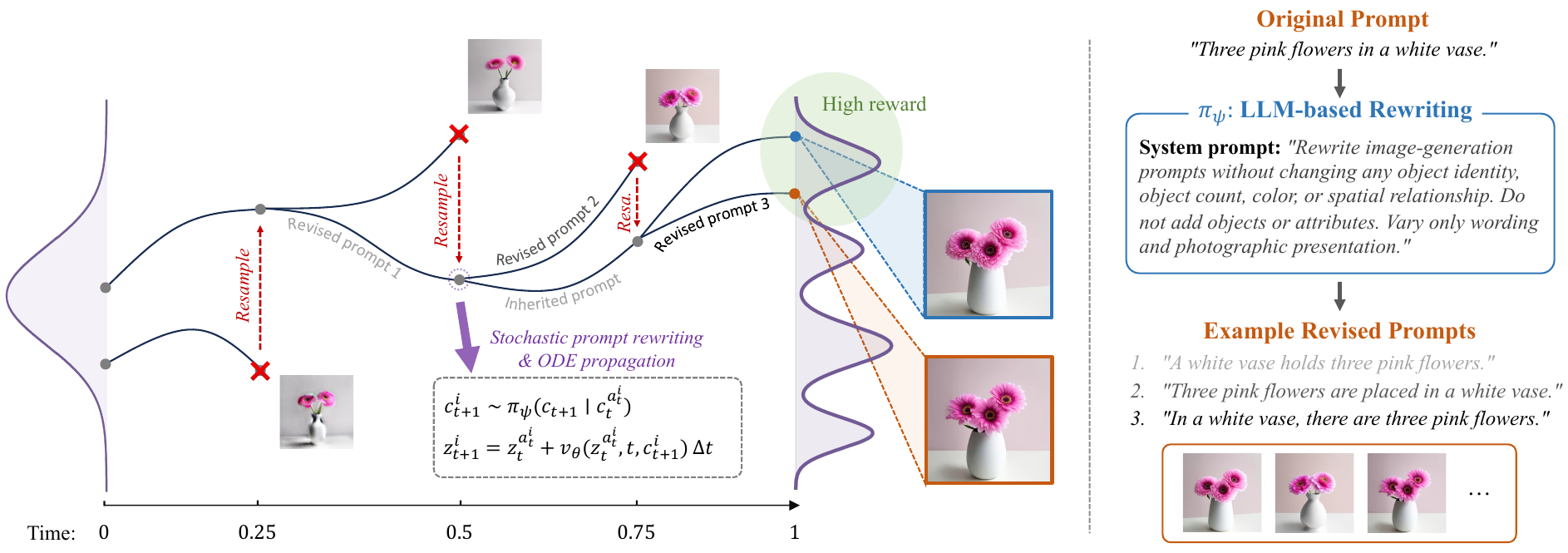}}
    \caption{Overview of our framework. \textbf{Left:} Reward-guided resampling selects promising particles, while LLM-based prompt rejuvenation enables duplicated particles to explore distinct continuations from the same latent state. Latent propagation follows a deterministic flow conditioned on each particle's revised prompt. \textbf{Right:} The rewriting model produces alternative phrasings under instructions to preserve object identities, counts, colors, and spatial relationships, introducing diversity through textual conditioning.
}
\label{fig:overview}
  \end{center}
  \vskip -0.2in
\end{figure*}

\section{Related Work}
\label{sec:relate}

\subsection{Text-to-Image Generation}

Diffusion models~\cite{ho2020denoising,song2020score} have been widely adopted for text-to-image generation, with models such as Imagen~\cite{saharia2022photorealistic} and latent diffusion~\cite{rombach2022high} demonstrating high-quality image synthesis conditioned on natural-language prompts. More recently, flow matching~\cite{lipman2022flow,liu2022flow} models offer a simple yet effective training formulation based on regressing velocity fields. They are ODE-based samplers, producing deterministic trajectories for a fixed initial noise and text condition. This formulation has been increasingly adopted in text-to-image models~\cite{esser2024scaling,labs2025flux,cai2025z,xie2024sana,qin2025lumina,cai2025hidream,wu2025qwen,zhao2026qwen}. Despite advances in visual quality, faithfully following complex prompts remains challenging, particularly when dealing with object counts, attribute bindings, or spatial relationships~\cite{ghosh2023geneval,huang2023t2i,ma2025janusflow}. These limitations motivate the development of new inference-time methods to achieve better alignment with original user prompts.

\subsection{Inference-Time Scaling}

To improve prompt-image alignment, inference-time scaling methods guide pretrained generators using reward feedback~\cite{uehara2025inference,Susladkar_2026_CVPR}. Commonly used strategies include gradient-based noise optimization~\cite{eyring2024reno,ben2024d}, verifier-guided search~\cite{ma2025inference,zhuo2025reflection,jiang2026raise}, and search over partial trajectories~\cite{li2025dynamic,jain2026diffusion,ramesh2026test}. 
On the other hand, particle-based approaches steer generation by maintaining multiple generation processes, weighting them using intermediate reward estimates, and resampling promising trajectories~\cite{wu2023practical,singhal2025general,zhang2026inference}. 
However, when combining with deterministic flow matching models, particles can be duplicated, as they follow from identical continuations. 
As such, it is often motivated to introduce stochastic dynamics or latent-state perturbations to restore diversity~\cite{kim2026inference}. Recent work also improves text-to-image generation through discrete prompt optimization~\cite{manas2024improving,khan2025test,kim2025reward,lee2026promptloop,kim2026rethinking,wu2026reprompt} or continuous embedding adaptation~\cite{na2026diffusion,kim2026test}. In this paper, our method introduces stochasticity through LLM-based prompt revision, enabling exploration of alternative trajectories while retaining deterministic ODE updates.

\section{Method}
\label{sec:method}

The proposed PromptSMC combines LLM-based prompt rejuvenation with reward-guided particle selection to improve prompt-image alignment, as illustrated in~\cref{fig:overview}.
In this section, we describe the exploration challenge in deterministic flow sampling, then introduce prompt rejuvenation and its integration into the sampling process.

\subsection{Problem Setup}
\label{sec:problem_setup}

Given a text prompt $c$, our goal is to generate images that align with $c$ using a flow matching model. Here, we consider latent-space generation, where a pretrained decoder $D_{\phi}$ maps a latent $z$ to an image $x=D_{\phi}(z)$. The flow matching model generates the latent state according to
\begin{equation}
    z_1=z_0+\int_0^1 v_{\theta}(z_t,t,c)\,\mathrm{d}t,
    \qquad z_0\sim\mathcal{N}(0,I),
    \label{eq:fm_generation}
\end{equation}
where $v_{\theta}(z_t,t,c)$ is the learned prompt-conditional velocity field at state $z_t$, with model parameter $\theta$. Sampling proceeds from source noise $z_0$ at $t=0$ to the terminal latent $z_1$ at $t=1$, which is then decoded into an image as $x=D_{\phi}(z_1)$. 
For notational simplicity, we use $t+1$ to denote the next sampling timestep in the discretized updates, with $\Delta t$ denoting the step size.
For an Euler solver, the update is
\begin{equation}
    z_{t+1}=z_t+ 
    v_{\theta}(z_t,t,c) \, \Delta t.
    \label{eq:deterministic_update}
\end{equation}

To formalize reward-guided generation, let $p_{\theta}(z_1\mid c)$ denote the terminal latent distribution induced by the pretrained flow model under a fixed prompt $c$. Given a reward model $r(x,c)$ that measures the alignment between prompts and generated images, the reward-tilted distribution~\cite{uehara2025inference} is
\begin{equation}
    p_r(z_1\mid c)
    \propto p_{\theta}(z_1\mid c)
    \exp(\lambda\,r(D_{\phi}(z_1),c)),
    \label{eq:reward_tilt}
\end{equation}
where $\lambda\geq 0$ controls the strength of reward guidance. This distribution formalizes the goal of favoring outputs that better align with the original prompt, motivating the use of sequential Monte Carlo (SMC), which maintains multiple interacting trajectories (i.e., particles) and progressively steers generation toward higher-reward outputs.
Since sampling directly from this distribution is computationally hard, our method uses intermediate rewards to guide particle selection while dynamically adapting the textual conditions during generation.

However, under the deterministic flow update in Eq.~\eqref{eq:deterministic_update}, resampling a particle into multiple copies does not create new continuations when their textual conditions remain unchanged. These copies follow the same subsequent trajectory, leading to identical outputs and limited particle exploration. 

To enable effective exploration after resampling, we maintain $N$ particles, each associated with a latent state $z_t^i$ and a prompt $c_t^i$, for $i=1,\ldots,N$. Here, $c$ remains the original user prompt used for evaluation, while $c_t^i$ is the particle's current generation prompt. This distinction allows us to explore different textual conditions while evaluating all particles against the same original user prompt.

\begin{figure}[t]
  \begin{center}
    \centerline{\includegraphics[width=\columnwidth]{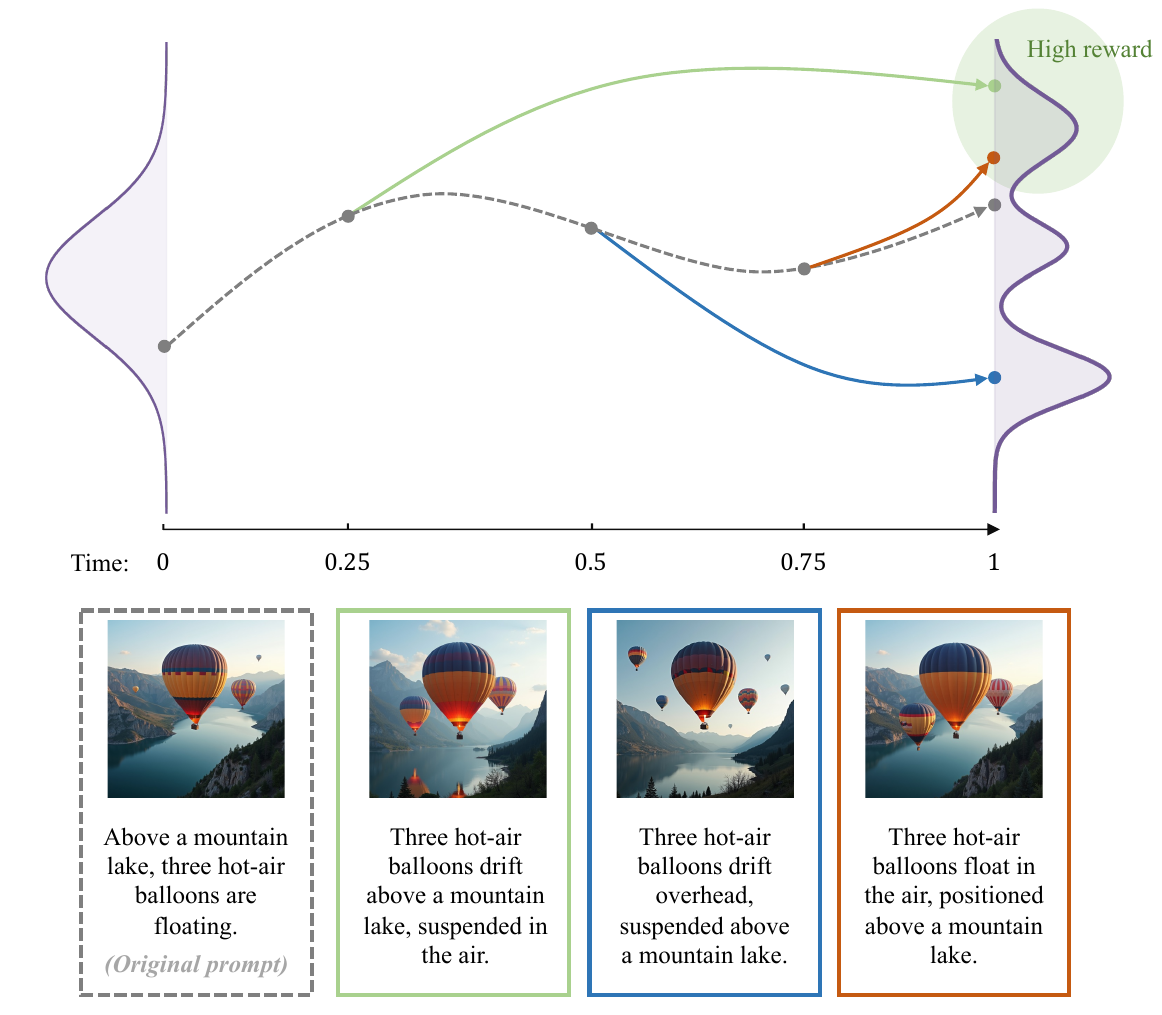}}
    \caption{LLM-based prompt rejuvenation for flow matching. Revising the prompt changes the conditional velocity field, enabling alternative trajectories without perturbing the state. 
    }
    \label{fig:prompt_rejuvenation}
  \end{center}
  \vskip -0.2in
\end{figure}

\subsection{LLM-Based Prompt Rejuvenation}
\label{sec:prompt_rejuvenation}

Since the velocity field $v_{\theta}$ depends on the text condition, revising the corresponding prompt can change the direction of subsequent generation, as illustrated in~\cref{fig:prompt_rejuvenation}. We therefore use this property to introduce alternative continuations for duplicated particles after resampling without explicitly perturbing their latent states.

Let $a_t^i$ denote the parent index selected for descendant~$i$ during resampling. Normally, the descendant inherits the parent's latent and prompt, i.e., $(z_t^{a_t^i}, c_t^{a_t^i})$. We improve this by leveraging a pretrained LLM to revise the prompt for each duplicated particle under semantic-preservation constraints. Denoting the LLM-based prompt rewriting procedure by $\pi_{\psi}$, we define the SMC propagate step as
\begin{align}
    c_{t+1}^i &\sim \pi_{\psi}(c_{t+1}\mid c_t^{a_t^i}), \label{eq:prompt_smc_update} \\
    z_{t+1}^i &= z_t^{a_t^i} + v_\theta(z_t^{a_t^i}, t, c_{t+1}^i) \, \Delta t.
 \label{eq:fm_smc_update} 
\end{align}
Here, Eq.~\eqref{eq:prompt_smc_update} denotes an LLM sampling process for prompt revision which introduces stochasticity into particles, while Eq.~\eqref{eq:fm_smc_update} is the flow matching deterministic update. More specifically, the LLM receives the inherited prompt and is instructed to preserve its objects, counts, colors, and spatial relationships while varying wording and presentation, as shown in the right of \cref{fig:overview}. These instructions encourage preservation of the original prompt's meaning. Intermediate images and reward scores are not provided to the LLM.

Although the descendants inherit the same latent state, their sampled prompts can induce different velocity predictions and lead them along different trajectories. We refer to this mechanism as \emph{prompt rejuvenation}, as it diversifies duplicated particles to explore alternative trajectories through prompt revision. The resulting exploration is driven by stochastic prompt generation, while ODE updates remain deterministic under each sampled prompt.

\begin{algorithm}[tb]
\caption{PromptSMC for Flow Matching}
\label{alg:promptsmc}
\begin{algorithmic}[1]
\REQUIRE Prompt $c$, flow matching model
$p_\theta(z_t \mid z_{t-1},c)$,
reward model $r(x,c)$, number of particles $N$

\STATE \textbf{Initialize:}
$z_0^i \sim \mathcal{N}(0,I)$,
$c_0^i = c$,
$w_0^i = 1/N$

\FOR{$t=0,\ldots,T-1$}
    \STATE \textbf{Resample:}
    Draw indices $a_t^i \sim \mathrm{Multinomial}(w_t)$
    and form the particle pairs
    $\{(z_t^{a_t^i},c_t^{a_t^i})\}_{i=1}^N$.

    \STATE \textbf{Propagate:}
    \STATE \quad
    $c_{t+1}^i \sim
    \pi_{\psi}(c_{t+1}\mid c_t^{a_t^i})$
    \STATE \quad
    $z_{t+1}^i = z_t^{a_t^i} + v_\theta(z_t^{a_t^i}, t, c_{t+1}^i) \, \Delta t$

    \STATE \textbf{Re-weight:}
    Compute
    $w_{t+1}^i =
    \exp\!\left(\lambda \cdot r(\hat{x}^i,c)\right)$
    and normalize the weights to sum to one.
\ENDFOR

\STATE \textbf{Output:} Samples $\{x^i\}_{i=1}^N$
\end{algorithmic}
\end{algorithm}

\begin{table*}
\caption{Quantitative comparison on the GenEval~\cite{ghosh2023geneval} benchmark across different backbones. $^\dagger$ denotes the reward used for particle guidance. GenEval is the main comparison metric. Best and second-best results are shown in \textbf{bold} and \underline{underline}, respectively.}
\vspace{-0.1in}
\label{tab:geneval_main}
\centering

\resizebox{\textwidth}{!}{
\begin{tabular}{
@{}l
c>{\columncolor{genevalgreen}}ccc
c>{\columncolor{genevalgreen}}ccc
c>{\columncolor{genevalgreen}}ccc
@{}
}
\toprule
& \multicolumn{4}{c}{SD3~\cite{esser2024scaling}}
& \multicolumn{4}{c}{FLUX.1-dev~\cite{labs2025flux}}
& \multicolumn{4}{c}{Z-Image~\cite{cai2025z}} \\
\cmidrule(lr){2-5}
\cmidrule(lr){6-9}
\cmidrule(lr){10-13}

Method
& ImageReward$^{\dagger}$ & GenEval & B-CLIP & VQAScore
& ImageReward$^{\dagger}$ & GenEval & B-CLIP & VQAScore
& ImageReward$^{\dagger}$ & GenEval & B-CLIP & VQAScore \\
\midrule

Vanilla
& 1.0674 & 0.7214 & 0.7885 & 0.8886
& 0.9588 & 0.6526 & 0.7797 & 0.8105
& 1.0790 & 0.7534 & 0.7957 & 0.9089 \\

FK (flow-adapted)~\cite{singhal2025general}
& 1.0906 & 0.6258 & 0.7846 & 0.7310
& 0.9423 & 0.4764 & 0.7722 & 0.6036
& 0.9690 & 0.6081 & 0.7860 & 0.7941 \\

DATE~\cite{na2026diffusion}
& 1.1397 & 0.7365 & 0.7872 & 0.8868
& 0.9863 & 0.6490 & 0.7805 & 0.7912
& 1.0892 & 0.7538 & 0.7934 & 0.9104 \\

SMC~\cite{singhal2025general}
& \underline{1.2896} & \underline{0.7568} & 0.7943 & 0.8988

& 1.2367 & \underline{0.6983} & 0.7876 & 0.8379

& 1.2688 & \underline{0.7673} & \underline{0.7975} & \textbf{0.9270} \\

VP-SDE+RBF~\cite{kim2026inference}
& 1.0984 & 0.6539 & 0.7849 & 0.8627
& 1.1051 & 0.6608 & 0.7822 & 0.8039
& 1.1445 & 0.6855 & 0.7885 & 0.8946 \\

VP-SDE+SMC~\cite{kim2026inference}
& 1.2802 & 0.7552 & 0.7956 & \underline{0.9001}
& 1.1962 & 0.6952 & 0.7816 & \underline{0.8442}
& 1.2487 & 0.7527 & 0.7932 & \underline{0.9204} \\

NR-SMC~\cite{lian2026testtime}
& 1.2512 & 0.7446 & \underline{0.7981} & 0.8928

& \underline{1.2485} & 0.6981 & \underline{0.7899} & 0.8359

& \underline{1.2754} & 0.7640 & 0.7963 & 0.9097 \\
\midrule

PromptSMC (Ours)
& \textbf{1.3331} & \textbf{0.7899} & \textbf{0.8026} & \textbf{0.9038}
& \textbf{1.3291} & \textbf{0.7699} & \textbf{0.7976} & \textbf{0.8783}

& \textbf{1.3083} & \textbf{0.8358} & \textbf{0.8041} & 0.9166 \\

\bottomrule
\end{tabular}
}
\end{table*}

\begin{table*}[t]
\centering
\small
\caption{Quantitative comparison on the T2I-CompBench~\cite{huang2023t2i} benchmark across different backbones. $^\dagger$ denotes the reward used for particle guidance. B-VQA is the main comparison metric. Best and second-best results are shown in \textbf{bold} and \underline{underline}, respectively.}
\vspace{-0.1in}
\label{tab:t2i_attribute_main}

\resizebox{\textwidth}{!}{
\begin{tabular}{
@{}l
c>{\columncolor{genevalgreen}}ccc
c>{\columncolor{genevalgreen}}ccc
c>{\columncolor{genevalgreen}}ccc
@{}
}
\toprule
& \multicolumn{4}{c}{SD3~\cite{esser2024scaling}}
& \multicolumn{4}{c}{FLUX.1-dev~\cite{labs2025flux}}
& \multicolumn{4}{c}{Z-Image~\cite{cai2025z}} \\
\cmidrule(lr){2-5}
\cmidrule(lr){6-9}
\cmidrule(lr){10-13}

Method
& ImageReward$^\dagger$
& B-VQA
& B-CLIP
& VQAScore
& ImageReward$^\dagger$
& B-VQA
& B-CLIP
& VQAScore
& ImageReward$^\dagger$
& B-VQA
& B-CLIP
& VQAScore \\
\midrule

Vanilla
& 1.1785 & 0.7111 & 0.7704 & 0.8673
& 0.9664 & 0.6300 & 0.7582 & 0.8227
& 1.1344 & 0.7292 & 0.7747 & 0.8766 \\

FK (flow-adapted)~\cite{singhal2025general}
& 1.0626 & 0.6793 & 0.7519 & 0.8524
& 0.7797 & 0.5445 & 0.7386 & 0.7907
& 0.8937 & 0.6455 & 0.7491 & 0.8558 \\

DATE~\cite{na2026diffusion}
& 1.2259 & 0.7153 & 0.7691 & 0.8619
& 1.0489 & 0.6380 & 0.7550 & 0.8279
& 1.1575 & 0.7310 & 0.7746 & 0.8804 \\

SMC~\cite{singhal2025general}
& 1.3784 & 0.7346 & 0.7735 & \underline{0.8803}

& 1.2653 & \underline{0.6723}
& \underline{0.7646} & \underline{0.8560}

& \underline{1.3305} & 0.7431
& \underline{0.7812} & \textbf{0.8931} \\

VP-SDE+RBF~\cite{kim2026inference}
& 1.1596 & 0.6902 & 0.7630 & 0.8410
& 1.0536 & 0.6386 & 0.7558 & 0.8236
& 1.1787 & 0.7201 & 0.7610 & 0.8606 \\

VP-SDE+SMC~\cite{kim2026inference}
& 1.3511 & \underline{0.7348} & \underline{0.7753} & 0.8707
& 1.1846 & 0.6631 & 0.7586 & 0.8417
& 1.2847 & \textbf{0.7568} & 0.7699 & 0.8843 \\

NR-SMC~\cite{lian2026testtime}
& \textbf{1.4015} & 0.7337 & 0.7741 & 0.8801
& \underline{1.2766} & 0.6701 & 0.7636 & 0.8548
& 1.3165 & 0.7398 & 0.7788 & 0.8859 \\

\midrule

PromptSMC (Ours)
& \underline{1.3854}
& \textbf{0.7407}
& \textbf{0.7790}
& \textbf{0.8809}
& \textbf{1.3054}
& \textbf{0.6887}
& \textbf{0.7726}
& \textbf{0.8701}

& \textbf{1.3535}
& \underline{0.7503}
& \textbf{0.7824}
& \underline{0.8884} \\

\bottomrule
\end{tabular}
}
\end{table*}

\subsection{Reward-Guided Sampling with PromptSMC}
\label{sec:prompt_smc_sampling}

Algorithm~\ref{alg:promptsmc} summarizes PromptSMC for general text-to-image flow matching models. Given an original prompt~$c$, we initialize $N$ particles with independent Gaussian latent states $z_0^i\sim\mathcal{N}(0,I)$, prompts $c_0^i=c$, and uniform weights $w_0^i=1/N$. Each particle maintains a latent state and a generation prompt, while the original prompt $c$ serves as the fixed reference for reward evaluation.

At each resampling step, particles are randomly selected according to their current weights, allocating more descendants to more promising trajectories. We then apply prompt rejuvenation to the duplicated particles, as described in the Sec.~\ref{sec:prompt_rejuvenation}. The descendants continue from their selected latent states using deterministic flow updates conditioned on the resulting prompts. This combines reward-guided selection with exploration through stochastic prompt revision.
To assess the updated particles before generation is complete, we estimate their terminal latent states using
\begin{equation}
    \hat{z}_{1 \mid t}^{i} = z_t^{i} + (1-t)\, v_\theta (z_t^{i}, t, c_t^{i})
    \label{eq:endpoint_estimate}
\end{equation}
and decode these estimates into images $\hat{x}_t^i= D_\phi(\hat{z}_{1 \mid t}^{i})$. A predefined reward model evaluates each estimate against the original prompt through $r(\hat{x}_t^i,c)$. The resulting scores determine the particle weights for subsequent resampling. 
All particles follow deterministic ODE updates between scheduled timesteps. 
At the end of sampling, we decode their terminal latent states into output images.

\section{Experiments}
\label{sec:exp}

\begin{figure}[t]
    \centering
    \includegraphics[width=\linewidth]
    {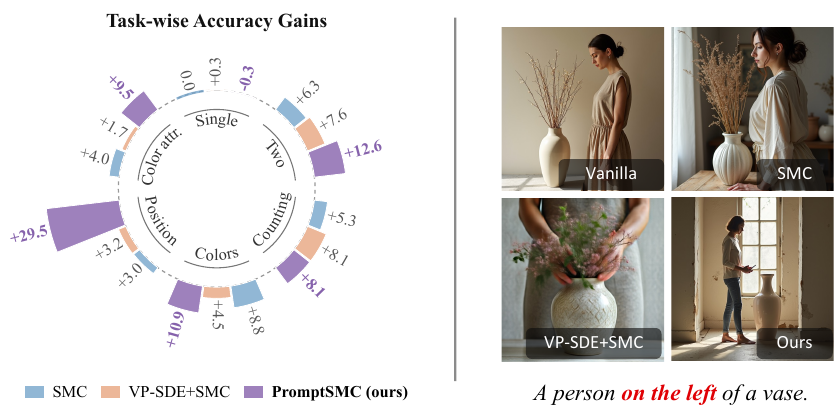}
    \caption{\textbf{Left:} Task-wise accuracy gains (in percentage points) of SMC-based approaches over vanilla FLUX.1-dev on GenEval. \textbf{Right:} Example of the position task.}
    \vspace{-0.1in}
    \label{fig:geneval_task_gains}
\end{figure}

\begin{figure*}[t]
    \centering
    \includegraphics[width=\linewidth]{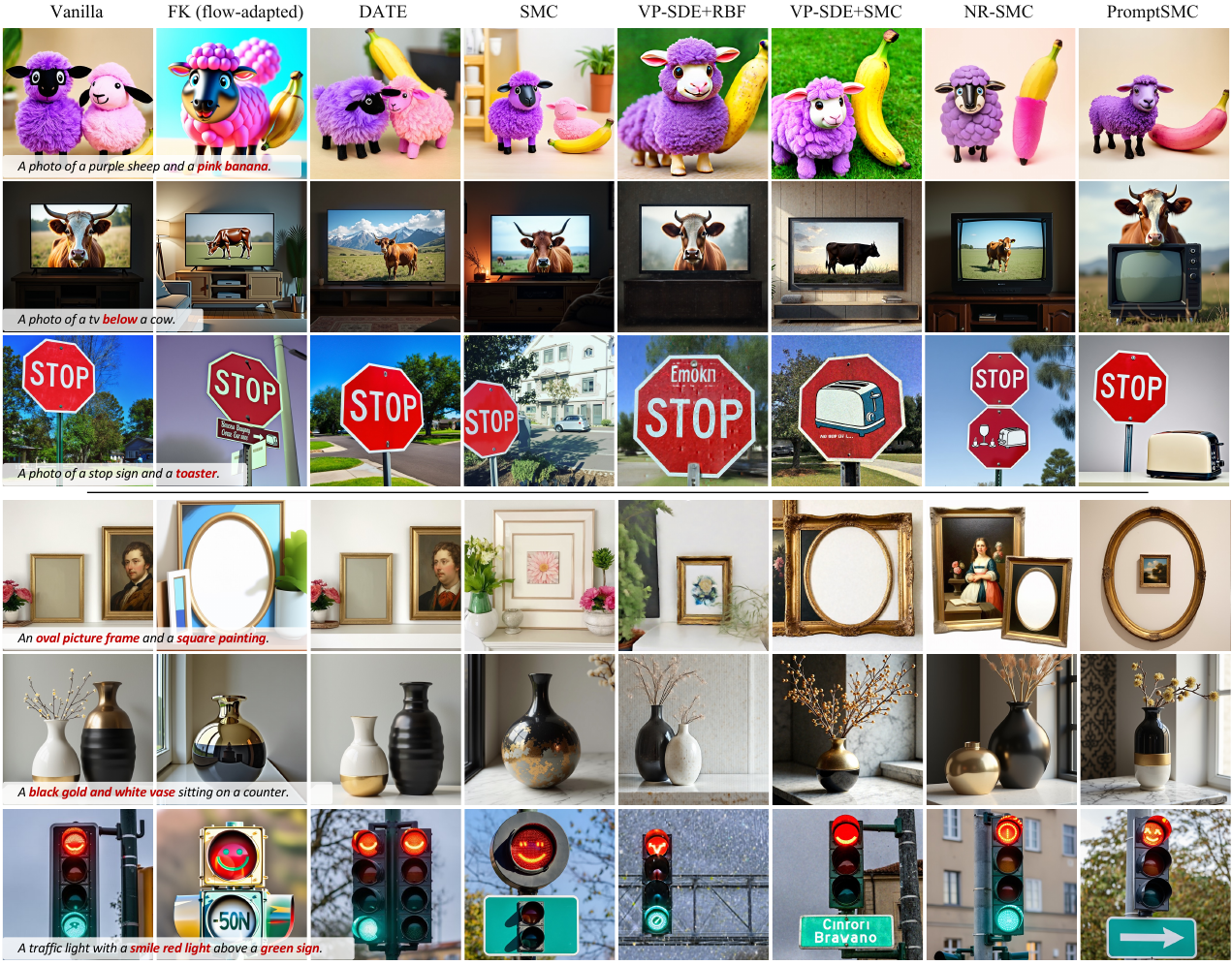}
    \caption{Visual comparison of our method with other state-of-the-art inference-time scaling approaches on GenEval (top three rows) and T2I-CompBench (bottom three rows). For each benchmark, the three rows show results from SD3, FLUX.1-dev, and Z-Image, respectively.}
    \label{fig:visual_results}
\end{figure*}

\subsection{Setup}
\label{sec:experimental_setup}

\paragraph{Benchmarks and Metrics.}
We evaluate our method on the GenEval~\cite{ghosh2023geneval} and T2I-CompBench~\cite{huang2023t2i} benchmarks. GenEval contains 553 prompts covering six tasks: single object, two objects, counting, colors, position, and color attribution. We use the standard GenEval overall score as our primary evaluation metric, defined as the mean accuracy across its six tasks. On T2I-CompBench, we evaluate the attribute-binding tasks (color, shape, and texture), with 300 prompts per category. We report the average BLIP-VQA (B-VQA)~\cite{huang2023t2i} score as the main metric. For both benchmarks, we additionally report ImageReward~\cite{xu2023imagereward}, BLIP-CLIP (B-CLIP)~\cite{chefer2023attend}, and VQAScore~\cite{lin2024evaluating} for reference. In the main comparisons, we use ImageReward as the reward function to guide particle selection, while the other reported metrics are used only for evaluation. 

\paragraph{Baselines.}
We use three pretrained flow-matching backbones: Stable Diffusion 3 (SD3)~\cite{esser2024scaling}, FLUX.1-dev~\cite{labs2025flux}, and Z-Image~\cite{cai2025z}. For each backbone, we compare against standard generation (Vanilla) and six inference-time scaling baselines: flow-adapted FK steering~\cite{singhal2025general}, DATE~\cite{na2026diffusion}, standard SMC without prompt rewriting~\cite{singhal2025general}, VP-SDE with rollover budget forcing (VP-SDE+RBF)~\cite{kim2026inference}, VP-SDE with SMC (VP-SDE+SMC)~\cite{kim2026inference}, and noise rejuvenated SMC (NR-SMC)~\cite{lian2026testtime}. Each method generates four images per prompt, all of which are used to compute the final metrics.

\paragraph{Implementation.} 
For all experiments, we use 50 integration steps and generate images at a resolution of $1024\times1024$. For particle-based approaches, we use $N=4$ particles, with resampling applied every five steps between steps 10 and 40. The reward-guidance strength is fixed at $\lambda=10$. We use Qwen3-VL-8B-Instruct~\cite{bai2025qwen3} as the default prompt rewriter, with text-only inputs and stochastic decoding. All experiments are conducted on a single NVIDIA GH200 GPU with 96 GB of memory. More implementation details are provided in the supplementary material (\cref{sec:app_impl}).

\subsection{Comparison Results}
\label{sec:comparison_results}

\paragraph{GenEval.}
The main quantitative results on GenEval are summarized in~\cref{tab:geneval_main}. Our method substantially improves over the vanilla model and achieves the highest GenEval score across all three backbones. The flow-adapted FK baseline obtains the lowest scores in our experiments. Other SMC-based approaches generally improve over the vanilla baseline, while PromptSMC achieves further gains by combining reward-guided resampling with LLM-based prompt rejuvenation. In addition to the GenEval score, PromptSMC achieves the highest B-CLIP scores and competitive VQAScore results. The task-wise results in~\cref{fig:geneval_task_gains} further show clear improvements in position, object counts, and color attribution. These results demonstrate the effectiveness of LLM-based prompt rejuvenation in improving prompt-image alignment in reward-guided flow matching.

\paragraph{T2I-CompBench.}
We additionally evaluate compositional prompt-image alignment on the attribute-binding tasks of T2I-CompBench, as shown in~\cref{tab:t2i_attribute_main}. Across color, shape, and texture binding, PromptSMC achieves the highest average BLIP-VQA score on SD3 and FLUX.1-dev and ranks second on Z-Image. It also obtains the highest B-CLIP scores across all three backbones. Since only ImageReward is used for particle guidance, gains on the other metrics indicate that PromptSMC improves prompt-image alignment beyond the optimized reward.

\paragraph{Qualitative results.}
Qualitative comparisons on GenEval and T2I-CompBench are shown in~\cref{fig:visual_results}.
Our method better captures the objects, attributes, and spatial relationships specified in the prompts. For example, it correctly generates a purple sheep and a pink banana, and accurately places a television below a cow, whereas other baselines are confused by the attributes or depict the cow on the television screen. These examples further illustrate the improvements in prompt-image alignment. More qualitative results are provided in the supplementary material (\cref{sec:quality}).

\section{Analysis and Discussion}
\label{sec:diss}

\begin{figure}[t]
  \centering
  \includegraphics[width=0.8\linewidth]{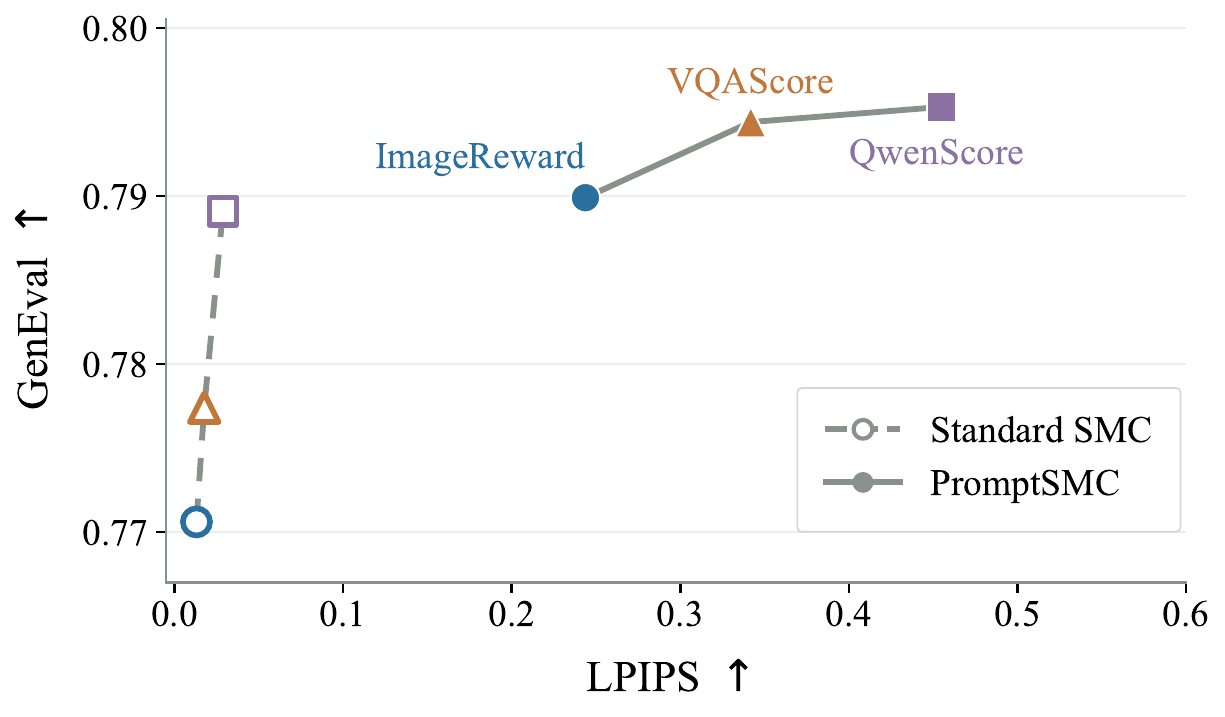}
  \vspace{-0.1in}
  \caption{Comparison of standard SMC and PromptSMC using different reward functions. Higher GenEval scores indicate better prompt-image alignment, while higher LPIPS indicates greater diversity among generated images per prompt.}
   \label{fig:reward_guidance}
\end{figure}

\begin{figure}[t]
  \centering
   \includegraphics[width=1.0\linewidth]{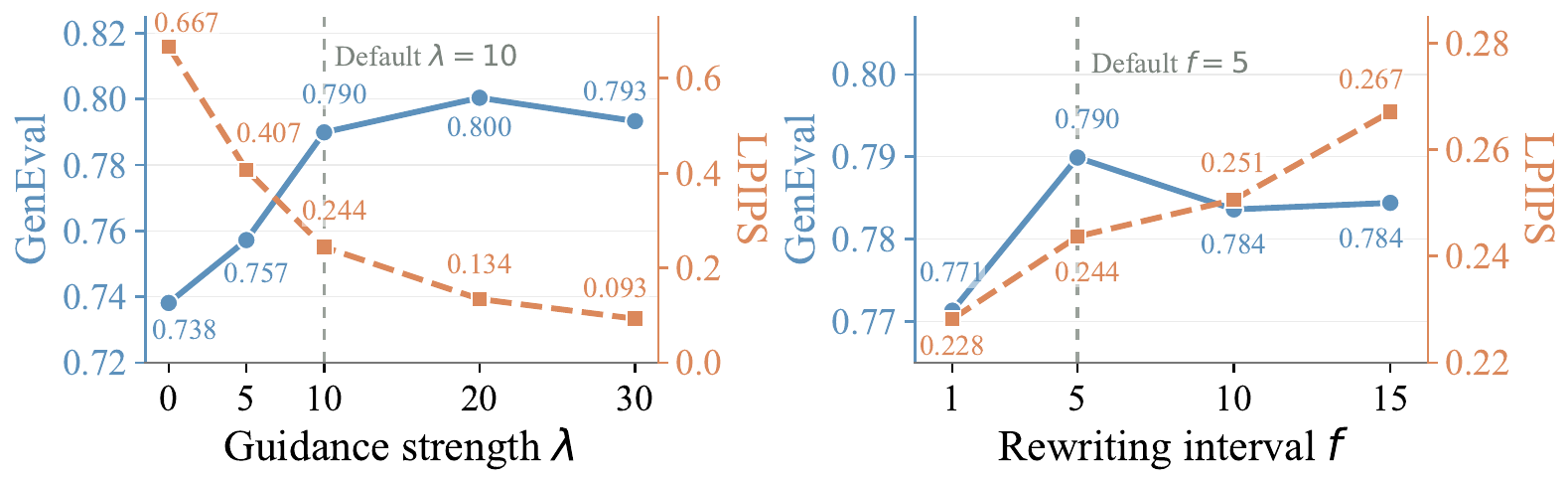}
    \vspace{-0.2in}
    \caption{Effect of reward guidance strength (\textbf{left}) and prompt rewriting interval (\textbf{right}) on GenEval using SD3.}
    \vspace{-0.1in}
    \label{fig:lambda_frequency}
\end{figure}

\subsection{Reward Guidance}

Following prior work~\cite{singhal2025general}, we use ImageReward~\cite{xu2023imagereward} as the default guidance objective for particle weighting and resampling in Sec.~\ref{sec:comparison_results}. To examine whether our method depends on a particular reward model, we also evaluate standard SMC and our PromptSMC using VQAScore~\cite{lin2024evaluating} and QwenScore (a variant of VQAScore using Qwen3-VL-8B-Instruct~\cite{bai2025qwen3}) as rewards, as shown in~\cref{fig:reward_guidance}. LPIPS~\cite{zhang2018unreasonable} measures the average pairwise diversity among generated images for each prompt. The results show that VQAScore and QwenScore can also provide effective guidance for both methods. Moreover, our PromptSMC achieves better prompt-image alignment while retaining greater sample diversity than standard SMC across different reward functions, demonstrating that its benefits are not specific to ImageReward. In addition, we further analyze the effect of reward guidance strength in~\cref{fig:lambda_frequency} (left). It shows that stronger guidance generally improves the GenEval score but can substantially degrade the diversity, revealing a trade-off between prompt-image alignment and sample diversity.

\begin{table}[t]
\centering
\small
\setlength{\tabcolsep}{6pt}
\caption{
Ablation study of prompt rejuvenation on GenEval using SD3. Vanilla (w/ rewriting) and SMC (w/ rewriting) apply prompt rewriting at initialization, using the same LLM as PromptSMC.
}
\vspace{-0.1in}
\label{tab:prompt_rejuvenation_ablation}
\resizebox{1.\columnwidth}{!}{
\begin{tabular}{@{}lcccc@{}}
\toprule
Method
& ImageReward
& GenEval
& Sec/prompt
& Memory (GB)\\
\midrule
Vanilla
& 1.0674 & 0.7214 & 14.10 & 35.01 \\

Vanilla (w/ rewriting)
& 1.0686 & 0.7381 & 21.31 & 52.55 \\

SMC 
& 1.2896 & 0.7568 & 18.35 & 36.81 \\

SMC (w/ rewriting)
& 1.3064 & 0.7590 & 21.91 & 54.34 \\

PromptSMC (Ours)
& \textbf{1.3331} & \textbf{0.7899} & 23.29 & 54.34\\
\bottomrule
\end{tabular}}
\end{table}

\begin{table}[t]
\caption{Comparison of different pretrained LLMs as prompt rewriters in PromptSMC on GenEval using SD3. Vanilla generation and SMC without prompt rewriting are included as references.}
\vspace{-0.1in}
\label{tab:rewriter_choice}
\centering
\small
\setlength{\tabcolsep}{3pt}

\resizebox{\columnwidth}{!}{
\begin{tabular}{@{}lcccc@{}}
\toprule
Method
& ImageReward
& GenEval

& Sec/prompt
& Memory (GB) \\
\midrule

Vanilla
& 1.0674
& 0.7214

& 14.10
& 35.01 \\

SMC (w/o rewriting)
& 1.2896
& 0.7568

& 18.35
& 36.81\\

\midrule

Ours (Gemma-3-4B-IT~\cite{team2025gemma})
& 1.3165
& 0.7686
& 31.60
& 51.26 \\

Ours (Qwen3-VL-4B~\cite{bai2025qwen3})
& 1.3184
& 0.7625
& 24.02
& 45.69 \\

\midrule

Ours (Llama-3.1-8B~\cite{grattafiori2024llama})
& 1.3185	
& 0.7744	
& 23.43	
& 52.90 \\

Ours (Qwen3-VL-8B~\cite{bai2025qwen3})
& \textbf{1.3331}
& \textbf{0.7899}
& 23.29 
& 54.34 \\

\bottomrule
\end{tabular}
}
\end{table}

\subsection{Prompt Rejuvenation}

As shown in~\cref{tab:prompt_rejuvenation_ablation}, applying prompt rewriting at initialization improves both Vanilla and SMC, while our PromptSMC further increases GenEval through dynamic prompt rejuvenation during sampling.
Notably, our method with rewriting interval $f=5$ is more effective than rewriting at every sampling step ($f=1$), as illustrated in~\cref{fig:lambda_frequency} (right). In addition,~\cref{tab:rewriter_choice} shows that different pretrained LLMs improve GenEval over SMC without rewriting, with Qwen3-VL-8B achieving the best overall results. This demonstrates the robustness of PromptSMC across different prompt rewriters. 
\cref{tab:best_of_n_prompt_comparison} further shows that PromptSMC outperforms both Best-of-$N$ and its prompt rewriting variant, demonstrating the effectiveness of prompt rejuvenation during sampling.
Finally, \cref{tab:rewrite_strategy} shows that minimal rewriting produces many duplicate prompts, while extensive rewriting generates longer and more diverse prompts. Our constrained rewriting strategy achieves the best performance, highlighting the importance of balancing semantic preservation and sufficient variation in textual conditions.
In the supplementary (Sec.~\ref{subsec:dts}), we further show that prompt rejuvenation generalizes beyond SMC. 

\begin{table}[t]
\centering
\small
\caption{
Comparison with Best-of-$N$ and its prompt rewriting variant on GenEval using SD3. All methods use $N=4$ particles and ImageReward-based top-1 selection for evaluation.
}
\vspace{-0.1in}
\label{tab:best_of_n_prompt_comparison}
\resizebox{\columnwidth}{!}{%
\begin{tabular}{@{}lcccc@{}}
\toprule
Method
& ImageReward
& GenEval
& Sec/prompt
& Memory (GB) \\
\midrule
Best-of-$N$
& 1.3703 & 0.7702 & 14.10 & 36.81 \\
Best-of-$N$ (w/ rewriting)
& 1.3860 & 0.7837 & 16.13 & 52.55 \\
PromptSMC (Ours)
& \textbf{1.4068} & \textbf{0.8022} & 23.29 & 54.34 \\
\bottomrule
\end{tabular}
}
\end{table}

\begin{table}[t]
\centering
\caption{Effect of prompt-rewriting instructions on a 120-prompt GenEval subset (20 prompts for each task). `\textit{Duplicate Rate}' is the percentage of resulting prompts that are duplicates, and `\textit{Length}' represents the average prompt length in words.}
\vspace{-0.1in}
\label{tab:rewrite_strategy}

\resizebox{\columnwidth}{!}{
\begin{tabular}{@{}lcccc@{}}
\toprule
Rewriting Strategy
& ImageReward
& GenEval
& Duplicate Rate
& Length \\
\midrule

No Rewrite
& 1.2998
& 0.7271
& 100\%
& 7.95 \\

Minimal Rewrite
& 1.3469
& 0.7313
& 67.9\%
& 9.99 \\

Extensive Rewrite
& 1.3598
& 0.7583
& 0.52\%
& 45.48 \\

Constrained (Ours)
& \textbf{1.3684}
& \textbf{0.7875}
& 5.12\%
& 21.40 \\

\bottomrule
\end{tabular}
}
\end{table}

\begin{figure}[t]
  \centering
   \includegraphics[width=1.0\linewidth]{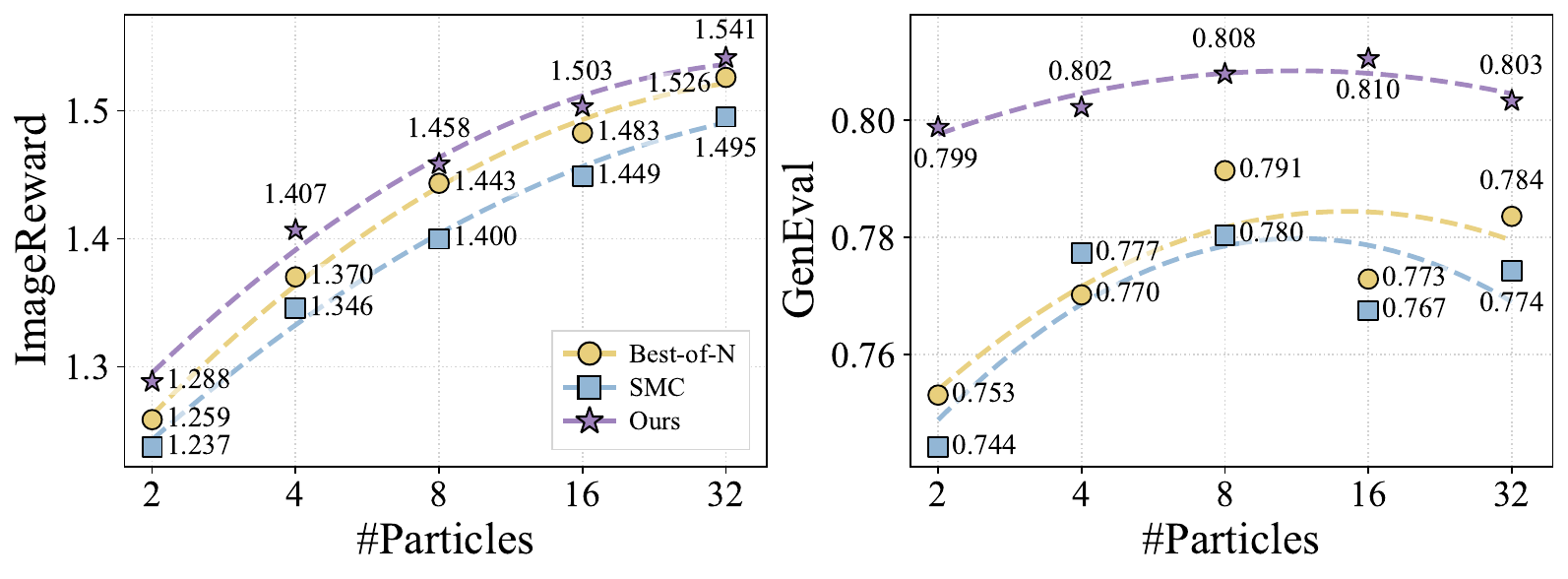}
    \vspace{-0.2in}
   \caption{Scaling the number of particles. All methods use ImageReward as guidance for particle selection.}
   \label{fig:particle_scaling}
\end{figure}

\subsection{Scaling the Number of Particles}

We study how PromptSMC scales with the particle number $N$ and compare it with Best-of-$N$ and standard SMC under the same particle budgets. All methods use ImageReward-based top-1 selection for evaluation. As shown in~\cref{fig:particle_scaling}, increasing $N$ generally improves ImageReward for all methods, while PromptSMC consistently achieves the highest ImageReward and GenEval scores at every evaluated $N$. Moreover, PromptSMC improves GenEval as the particle number increases from $N=2$ to $16$, whereas Best-of-$N$ and standard SMC exhibit less consistent scaling. These results indicate that increasing the number of particles brings greater benefits when prompt rejuvenation is incorporated into particle evolution.
At $N=32$, our method further improves ImageReward but slightly decreases GenEval, suggesting that performance eventually saturates as the particle number grows. In the main experiments, we use $N=4$ as it provides a favorable balance between generation performance and computational cost.

\subsection{Performance-Efficiency Trade-off}

We first compare different configurations of particle counts and sampling steps under a fixed NFE budget. As shown in~\cref{tab:compute_comparison}, increasing the number of sampling steps for vanilla generation leads to only limited improvements, even with substantially more computation. Under the same NFE budget, standard SMC outperforms vanilla generation, while PromptSMC further improves both ImageReward and GenEval through prompt-based particle exploration. Among the tested configurations, using 4 particles with 50 sampling steps achieves the highest GenEval score while requiring less runtime than configurations with more particles.
\cref{tab:compute_cost} compares the runtime and memory requirements of different inference-time scaling methods. PromptSMC achieves the highest ImageReward and GenEval results at the cost of additional runtime and memory for LLM-based prompt rejuvenation. Overall, PromptSMC achieves better prompt-image alignment with reasonable computational overhead.

\begin{table}[t]
\centering
\small
\caption{Comparison of different particle counts and sampling steps on GenEval using SD3. All methods use 200 model function evaluations, i.e., $N \times \mathrm{Steps}$, except for the Vanilla-long baseline.}
\vspace{-0.1in}
\label{tab:compute_comparison}
\resizebox{\columnwidth}{!}{
\begin{tabular}{lcccccc}
\toprule
Method
& $N$
& Steps
& ImageReward
& GenEval
& Sec/prompt
& Memory (GB) \\
\midrule

Vanilla
& 4
& 50
& 1.0674
& 0.7214
& 14.10	
& 35.01 \\

Vanilla-long
& 4
& 200
& 1.0951
& 0.7355
& 56.02
& 35.01 \\

SMC
& 4
& 50
& 1.2896
& 0.7568
& 18.35 
& 36.81 \\

\midrule

PromptSMC
& 10
& 20
& 1.3370
& 0.7830
& 41.49
& 54.47 \\

PromptSMC
& 20
& 10
& 1.3025
& 0.7840
& 82.63
& 54.99 \\

PromptSMC
& 4
& 50
& 1.3331
& 0.7899
& 23.29
& 54.34 \\

\bottomrule
\end{tabular}}
\end{table}

\begin{table}[t]
\centering
\small
\setlength{\tabcolsep}{4pt}
\caption{Performance and computational efficiency comparison on GenEval using SD3.}
\vspace{-0.1in}
\label{tab:compute_cost}

\resizebox{\columnwidth}{!}{
\begin{tabular}{lcccc}
\toprule
Method
& ImageReward
& GenEval
& Sec/prompt
& Memory (GB)\\
\midrule

Vanilla
& 1.0674 
& 0.7214 
& 14.10
& 35.01 \\

FK (flow-adapted)~\cite{singhal2025general}
& 1.0906
& 0.6258
& 18.28
& 36.81 \\

DATE~\cite{na2026diffusion}
& 1.1397
& 0.7365
& 106.7
& 44.98\\

SMC~\cite{singhal2025general}
& 1.2896
& 0.7568
& 18.35
& 36.81 \\

VP-SDE+RBF~\cite{kim2026inference}
& 1.0984 
& 0.6539 
& 56.05
& 36.81 \\

VP-SDE+SMC~\cite{kim2026inference}
& 1.2802
& 0.7552
& 18.72
& 36.81 \\

NR-SMC~\cite{lian2026testtime}
& 1.2512
& 0.7446
& 20.26
& 36.81 \\

PromptSMC (Ours)
& \textbf{1.3331}
& \textbf{0.7899}
& 23.29
& 54.34 \\

\bottomrule
\end{tabular}}
\end{table}

\section{Conclusion}
This work introduced an LLM-based prompt rejuvenation approach for particle exploration in text-to-image flow matching. The proposed method, PromptSMC, uses a pretrained LLM to stochastically revise textual conditions while retaining intermediate latent states, enabling alternative trajectory exploration under deterministic ODE updates. PromptSMC leverages prompt rejuvenation with reward-guided particle selection to improve prompt-image alignment, without relying on noise injection, model training, or reward gradients. 
Experiments across multiple benchmarks and flow matching backbones demonstrate consistent improvements in prompt-image alignment, showing that prompt-based exploration is an effective mechanism for particle-based text-to-image generation.

\section*{Acknowledgments}
This research was supported by \emph{Kjell och M{\"a}rta Beijer Foundation}. 
Computational resources were provided by the National Academic Infrastructure for Supercomputing in Sweden (NAISS), funded by the Swedish Research Council. 
We also thank Ziwei Luo and Wenjing Lian for their helpful feedback.

\section*{LLM Usage Statement}
In this project, ChatGPT (OpenAI) was used for grammar correction and language polishing, while Claude Code (Anthropic) was used for code review and typo checking. All technical content, experimental results, and conclusions were reviewed and verified by the authors.

{
    \small
    \bibliographystyle{ieeenat_fullname}
    \bibliography{main}
}

\clearpage
\setcounter{page}{12}
\maketitlesupplementary
\setcounter{table}{8}

\appendix
\section{Overview}

In this supplementary material, we provide:
\begin{itemize}
    \item In~\cref{sec:app_impl}, we provide additional implementation details and information about the baseline models and our PromptSMC used in our experiments.
    \item We explore an alternative prompt-rewriting strategy in our framework in \cref{subsec:template_prompt}. In~\cref{subsec:dts}, we further apply prompt rejuvenation to another particle-based method to evaluate its effectiveness beyond SMC.
    \item In~\cref{sec:app_quantity}, we present additional quantitative results. We report task-wise results on GenEval and T2I-CompBench for our method and all baselines in~\cref{subsec:task_wise}. We also present more comprehensive results on particle scaling, guidance strength, and resampling settings in \cref{subsec:app_particle} and \cref{subsec:app_resample}. We provide additional performance efficiency analysis in \cref{subsec:efficient}.
    \item \cref{sec:more_quality} provides additional qualitative results, including examples illustrating our motivation, comparisons with baselines, and further generation examples. We also present and analyze several challenging cases to further examine the limitations of our method in \cref{subsec:failure_cases}.
\end{itemize}

\section{Implementation Details}
\label{sec:app_impl}

\subsection{Backbones \& Baselines}
\label{subsec:app_backbones}

We evaluate our framework on three pretrained text-to-image flow matching models: Stable Diffusion 3 Medium (SD3)~\cite{esser2024scaling}, FLUX.1-dev~\cite{labs2025flux}, and Z-Image~\cite{cai2025z}. We use the released checkpoints listed in ~\cref{app-tab:app_backbones} without fine-tuning.
For baselines, DATE~\cite{na2026diffusion} adapts text embeddings during sampling using reward gradients. FK (flow-adapted)~\cite{singhal2025general} and standard SMC~\cite{singhal2025general} share the same reward-guided particle framework, but FK injects additional noise into the flow matching updates. VP-SDE+RBF~\cite{kim2026inference} combines variance-preserving SDE sampling with adaptive compute allocation across timesteps, while VP-SDE+SMC combines the same sampling process with SMC resampling. NR-SMC~\cite{lian2026testtime} introduces noise-based particle rejuvenation to enable exploration after resampling.

\begin{table*}[t]
\caption{
Backbones used in our experiments. Parameter counts exclude auxiliary text encoders.}
\vspace{-0.1in}
\label{app-tab:app_backbones}
\centering
\small
\begin{tabularx}{\textwidth}{
@{}l >{\raggedright\arraybackslash}X r l @{}
}
\toprule
Backbone & Checkpoint & Params. & Variant  \\
\midrule
SD3 Medium
& \texttt{stabilityai/stable-diffusion-3-medium-diffusers}
& 2B & Medium  \\
FLUX
& \texttt{black-forest-labs/FLUX.1-dev}
& 12B & Guidance-distilled  \\
Z-Image
& \texttt{Tongyi-MAI/Z-Image}
& 6B & Undistilled base  \\
\bottomrule
\end{tabularx}
\end{table*}

\subsection{Prompt Rejuvenation}
\label{subsec:app_prompt_models}

We use \texttt{Qwen/Qwen3-VL-8B-Instruct}~\cite{bai2025qwen3}
as the default prompt rewriter.
Although the model supports visual inputs, we use it only for text-to-text prompt rewriting; generated images are not provided
to the rewriter in the main comparisons. We also consider other LLMs as prompt rewriters, as listed in~\cref{app-tab:prompt_rewriters}. We use the same system and user prompt templates for these models.
We format these messages using each model’s chat template.

The exact prompt templates are provided below. All experiments use the constrained system prompt, except for the rewriting-strategy ablation in~\cref{tab:rewrite_strategy}, where it is replaced by the minimal or extensive rewrite system prompt. The user prompt is shared across all settings, where \texttt{\{prompt\}} denotes the text passed to the rewriter at the current rewriting step. Note that Algorithm~\ref{alg:promptsmc} simply presents resampling and prompt rejuvenation at every sampling step for simplicity. In practice, these operations are applied only at scheduled steps, with deterministic ODE update performed at every step. By default, we check for resampling every five steps within the window $[10, 40)$. After resampling, for each selected parent, one descendant particle retains the inherited prompt, while other descendant particles will change their prompts using the proposed prompt rejuvenation. 

\begin{promptbox}{System prompt (Constrained)}
Rewrite image-generation prompts without changing any object identity, object count, color, or spatial relationship. Do not add objects or attributes. Vary only wording and photographic presentation.
\end{promptbox}

\begin{promptbox}{System prompt (Minimal Rewrite)}
Rephrase without changing meaning or adding details.
\end{promptbox}

\begin{promptbox}{System prompt (Extensive Rewrite)}
You are an expert at enhancing image generation prompts. Rewrite the user's prompt to be more detailed and descriptive while keeping the original intent. Make it clear, well-structured, and suitable for image generation.
\end{promptbox}

\begin{promptbox}{User prompt}
Rewrite this prompt with clearer visual wording: {prompt}
Do not invent a setting.
Return exactly {"revised_prompt": "..."}.
\end{promptbox}

\begin{table}[t]
\caption{
Overview of the LLMs used for prompt rewriting. All models are only used for text-to-text rewriting.
}
\vspace{-0.1in}
\label{app-tab:prompt_rewriters}
\centering
\resizebox{\columnwidth}{!}{
\begin{tabular}{@{}ll@{}}
\toprule
Model & Checkpoint \\
\midrule

Qwen3-VL-4B
& \texttt{Qwen/Qwen3-VL-4B-Instruct} \\

Gemma-3-4B-IT
& \texttt{google/gemma-3-4b-it} \\

\midrule

\rowcolor{defaultrow}
\textbf{Qwen3-VL-8B (default)}
& \texttt{Qwen/Qwen3-VL-8B-Instruct} \\

Llama-3.1-8B
& \texttt{meta-llama/Llama-3.1-8B-Instruct} \\

\bottomrule
\end{tabular}
}
\end{table}

\subsection{Reward Models and Evaluation}
\label{sec:app_reward_evaluation}

Unless otherwise specified, we use ImageReward-v1.0~\cite{xu2023imagereward} as our default reward model to guide intermediate particle selection. Here, ImageReward is used for both guidance and evaluation. To study the effect of the guidance reward, we investigate the standard VQAScore with CLIP-FlanT5-XXL~\cite{lin2024evaluating} and a variant using Qwen3-VL-8B-Instruct~\cite{bai2025qwen3} (denoted as QwenScore) as alternative reward functions.
In practice, the reweighting step in Algorithm~\ref{alg:promptsmc} uses a maximum-reward potential. At each scheduled check before the final check, we compute $\widetilde{w}_t^i = \exp\!\left(\lambda m_t^i\right)$ and normalize the weights across particles. Here, $m_t^i$ is the running maximum reward along particle's trajectory, initialized to zero and updated with the current reward.

\paragraph{Dual roles of vision-language models (VLMs).}
Notably, VLMs such as Qwen3-VL-8B-Instruct
can be used for both reward guidance and prompt rewriting. The reward model evaluates how well a generated image satisfies the original user prompt, whereas the prompt rewriter modifies the text condition used for subsequent generation.

\paragraph{Evaluation.}
In addition to the standard metrics provided by each benchmark, we also report B-CLIP~\cite{chefer2023attend} and VQAScore~\cite{lin2024evaluating} as additional post-hoc evaluation metrics in the default setting. B-CLIP measures the CLIP text similarity between the prompt and the BLIP-generated caption of the image, while VQAScore measures the probability that a VQA model answers ``Yes'' to whether the image depicts the prompt. 
For the main comparisons in~\cref{tab:geneval_main} and \cref{tab:t2i_attribute_main}, we generate four images per prompt and include all of them in the evaluation. Following GenEval~\cite{ghosh2023geneval}, we compute accuracy for each of the six tasks and report their mean as the overall score. For T2I-CompBench~\cite{huang2023t2i}, we report the mean B-VQA score across the color, shape, and texture categories. For the top-1 comparisons in \cref{tab:best_of_n_prompt_comparison} and \cref{fig:particle_scaling}, we score the $N$ final candidates against the original prompt using ImageReward. Only the highest-scoring image from each prompt contributes to the reported quality metrics. For LPIPS~\cite{zhang2018unreasonable}, we compute the mean of the six pairwise distances among the four images of each prompt, and then average over all prompts.

\paragraph{Runtime and memory measurement.}
All runtime and memory measurements are conducted on a single NVIDIA
GH200 GPU. Runtime is measured per prompt from prompt encoding to image
decoding, including denoising, intermediate reward evaluation (where
applicable), and prompt rewriting (for methods with prompt rewriting). It excludes model loading, final candidate scoring and selection, image saving, and metric computation. 
For GPU memory, we report the peak
memory allocated by PyTorch over all prompts (in GB). Note that
Best-of-$N$ has the same runtime as Vanilla generation with $N$
samples, but requires more GPU memory because the ImageReward model is
loaded for final candidate selection.

\begin{table}[t]
\centering
\caption{
Comparison of our method using template-based and LLM-based prompt rewriting on GenEval.
}
\vspace{-0.1in}
\label{app-tab:template}
\resizebox{\columnwidth}{!}{
\begin{tabular}{@{}llc>{\columncolor{genevalgreen}}ccc@{}}
\toprule
Backbone & Method
& ImageReward$^\dagger$ $\uparrow$
& GenEval $\uparrow$
& B-CLIP $\uparrow$
& VQAScore $\uparrow$ \\
\midrule

\multirow{4}{*}{SD3}
& Vanilla
& 1.0674 & 0.7214 & 0.7885 & 0.8886 \\

& SMC
& 1.2896 & 0.7568 & 0.7943 & 0.8988 \\

& Ours (Template)
& 1.2902 & 0.7653 & 0.7953 & 0.9011 \\

& Ours (LLM)
& \textbf{1.3331}
& \textbf{0.7899}
& \textbf{0.8026}
& \textbf{0.9038} \\

\midrule

\multirow{4}{*}{FLUX.1-dev}
& Vanilla
& 0.9588 & 0.6526 & 0.7797 & 0.8105 \\

& SMC
& 1.2367 & 0.6983 & 0.7876 & 0.8379 \\

& Ours (Template)
& 1.2838 & 0.7139 & 0.7892 & 0.8428 \\

& Ours (LLM)
& \textbf{1.3291}
& \textbf{0.7699}
& \textbf{0.7976}
& \textbf{0.8783} \\

\bottomrule
\end{tabular}
}
\end{table}

\begin{table}[t]
\centering
\caption{
Generalization of prompt rejuvenation to DTS on SD3. LPIPS measures the average pairwise diversity. NFE and runtime are reported per prompt.
Best values are shown in bold.
}
\vspace{-0.1in}
\label{tab:sd3_dts_rewriting}

\resizebox{\columnwidth}{!}{
\begin{tabular}{@{}lc>{\columncolor{genevalgreen}}cccc@{}}
\toprule
Method
& ImageReward $\uparrow$
& GenEval $\uparrow$
& NFE
& Sec/prompt $\downarrow$
& LPIPS $\uparrow$ \\
\midrule

DTS
& 1.1878
& 0.7415
& 228
& \textbf{17.17}
& 0.4026 \\

+ Prompt Rejuven.
& \textbf{1.2087}
& \textbf{0.7613}
& 228
& 21.31
& \textbf{0.4329} \\

\bottomrule
\end{tabular}
}
\end{table}

\begin{table*}[t]
\centering
\small

\begin{minipage}{\textwidth}
\centering
\small
\caption{
Task-wise accuracy on GenEval using SD3. Higher is better.
The overall GenEval column is shaded to highlight the benchmark metric.
The best and second-best results in each column are shown in
\textbf{bold} and \underline{underline}, respectively.
}
\label{app-tab:geneval_breakdown_sd3}

\begin{tabular*}{\linewidth}{
@{\extracolsep{\fill}}l>{\columncolor{genevalgreen}}cccccccc@{}
}
\toprule
& \multicolumn{1}{c}{} & & \multicolumn{2}{c}{Objects} & & & & \\
\cmidrule(lr){4-5}
Method & Overall & Correct Images & Single & Two
& Counting & Colors & Position & Color Attr. \\
\midrule
Vanilla~\cite{esser2024scaling}
& 0.7214 & 0.7107 & 0.9938 & 0.8838
& 0.7000 & 0.8457 & 0.2950 & 0.6100 \\

FK (flow-adapted)~\cite{singhal2025general}
& 0.6258 & 0.6130 & \underline{0.9969} & 0.8611
& 0.5688 & 0.7207 & 0.2550 & 0.3525 \\

DATE~\cite{na2026diffusion}
& 0.7365 & 0.7269 & 0.9906 & 0.9116
& 0.6969 & \textbf{0.8723} & 0.3150 & 0.6325 \\

SMC~\cite{singhal2025general}
& \underline{0.7568} & \underline{0.7464}
& 0.9844 & \underline{0.9293}
& 0.7750 & \underline{0.8670}
& 0.3425 & \underline{0.6425} \\

VP-SDE+RBF~\cite{kim2026inference}
& 0.6539 & 0.6401 & 0.9906 & 0.7500
& 0.6438 & 0.8138 & 0.2500 & 0.4750 \\

VP-SDE+SMC~\cite{kim2026inference}
& 0.7552 & 0.7432 & 0.9906 & 0.9066
& \textbf{0.8156} & 0.8457 & 0.3350 & 0.6375 \\

NR-SMC~\cite{lian2026testtime}
& 0.7446 & 0.7337 & 0.9938 & 0.9242
& 0.7688 & 0.8032 & \underline{0.3450} & 0.6325 \\
\midrule
PromptSMC (Ours)
& \textbf{0.7899} & \textbf{0.7816}
& \textbf{1.0000} & \textbf{0.9596}
& \underline{0.7781} & 0.8644
& \textbf{0.4175} & \textbf{0.7200} \\
\bottomrule
\end{tabular*}
\end{minipage}

\par\vspace{1em}

\begin{minipage}{\textwidth}
\centering
\small
\caption{
Task-wise accuracy on GenEval using FLUX.1-dev. Higher is better.
The overall GenEval column is shaded to highlight the benchmark metric.
The best and second-best results in each column are shown in
\textbf{bold} and \underline{underline}, respectively.
}
\label{app-tab:geneval_breakdown_flux}

\begin{tabular*}{\linewidth}{
@{\extracolsep{\fill}}l>{\columncolor{genevalgreen}}cccccccc@{}
}
\toprule
& \multicolumn{1}{c}{} & & \multicolumn{2}{c}{Objects} & & & & \\
\cmidrule(lr){4-5}
Method & Overall & Correct Images & Single & Two
& Counting & Colors & Position & Color Attr. \\
\midrule
Vanilla~\cite{labs2025flux}
& 0.6526 & 0.6365 & \underline{0.9906} & 0.7854
& 0.7219 & 0.7553 & 0.2000 & 0.4625 \\

FK (flow-adapted)~\cite{singhal2025general}
& 0.4764 & 0.4571 & 0.8844 & 0.6010
& 0.5969 & 0.4814 & 0.1375 & 0.1575 \\

DATE~\cite{na2026diffusion}
& 0.6490 & 0.6334 & 0.9875 & 0.7727
& 0.6938 & 0.7926 & 0.2000 & 0.4475 \\

SMC~\cite{singhal2025general}
& \underline{0.6983} & \underline{0.6831}
& \underline{0.9906} & 0.8485
& 0.7750 & \underline{0.8431}
& 0.2300 & \underline{0.5025} \\

VP-SDE+RBF~\cite{kim2026inference}
& 0.6608 & 0.6442 & \underline{0.9906} & 0.7904
& 0.7500 & 0.7739 & 0.2050 & 0.4550 \\

VP-SDE+SMC~\cite{kim2026inference}
& 0.6952 & 0.6790 & \textbf{0.9938} & \underline{0.8611}
& \textbf{0.8031} & 0.8005 & 0.2325 & 0.4800 \\

NR-SMC~\cite{lian2026testtime}
& 0.6981 & 0.6826 & \underline{0.9906} & 0.8535
& \underline{0.7938} & 0.8005 & \underline{0.2775} & 0.4725 \\
\midrule
PromptSMC (Ours)
& \textbf{0.7699} & \textbf{0.7595}
& 0.9875 & \textbf{0.9116}
& \textbf{0.8031} & \textbf{0.8644}
& \textbf{0.4950} & \textbf{0.5575} \\
\bottomrule
\end{tabular*}
\end{minipage}

\par\vspace{1em}

\begin{minipage}{\textwidth}
\centering
\small
\caption{
Task-wise accuracy on GenEval using Z-Image. Higher is better.
The overall GenEval column is shaded to highlight the benchmark metric.
The best and second-best results in each column are shown in
\textbf{bold} and \underline{underline}, respectively.
}
\label{app-tab:geneval_breakdown_zimage}

\begin{tabular*}{\linewidth}{
@{\extracolsep{\fill}}l>{\columncolor{genevalgreen}}cccccccc@{}
}
\toprule
& \multicolumn{1}{c}{} & & \multicolumn{2}{c}{Objects} & & & & \\
\cmidrule(lr){4-5}
Method & Overall & Correct Images & Single & Two
& Counting & Colors & Position & Color Attr. \\
\midrule
Vanilla~\cite{cai2025z}
& 0.7534 & 0.7428 & \underline{0.9969} & 0.9167
& 0.7656 & 0.8564 & 0.4075 & 0.5775 \\

FK (flow-adapted)~\cite{singhal2025general}
& 0.6081 & 0.5922 & 0.9625 & 0.8056
& 0.6906 & 0.5824 & 0.3600 & 0.2475 \\

DATE~\cite{na2026diffusion}
& 0.7538 & 0.7428 & \underline{0.9969} & 0.8990
& 0.7719 & 0.8750 & 0.4075 & 0.5725 \\

SMC~\cite{singhal2025general}
& \underline{0.7673} & \underline{0.7572}
& \underline{0.9969} & 0.9091
& 0.7656 & \textbf{0.9149}
& 0.4050 & \underline{0.6125} \\

VP-SDE+RBF~\cite{kim2026inference}
& 0.6855 & 0.6736 & 0.9844 & 0.8030
& 0.6750 & 0.7979 & 0.3750 & 0.4775 \\

VP-SDE+SMC~\cite{kim2026inference}
& 0.7527 & 0.7405 & 0.9875 & 0.9192
& \underline{0.8313} & 0.8059 & 0.4150 & 0.5575 \\

NR-SMC~\cite{lian2026testtime}
& 0.7640 & 0.7536 & \textbf{1.0000} & \underline{0.9419}
& 0.7813 & 0.8511 & \underline{0.4750} & 0.5350 \\
\midrule
PromptSMC (Ours)
& \textbf{0.8358} & \textbf{0.8287}
& \textbf{1.0000} & \textbf{0.9722}
& \textbf{0.8438} & \underline{0.8989}
& \textbf{0.6700} & \textbf{0.6300} \\
\bottomrule
\end{tabular*}
\end{minipage}

\end{table*}

\begin{table*}[t]
    \caption{
    Results on T2I-CompBench across three flow matching backbones. Overall is the arithmetic mean of the Color, Shape, and Texture B-VQA scores. The best and second-best results within each backbone are shown in \textbf{bold} and \underline{underline}, respectively.
    }
    \vspace{-0.1in}
    \label{tab:t2i_attribute_breakdown}
    \centering
    \resizebox{\textwidth}{!}{
    \begin{tabular}{@{}l>{\columncolor{genevalgreen}}cccc>{\columncolor{genevalgreen}}cccc>{\columncolor{genevalgreen}}cccc@{}}
    \toprule
    & \multicolumn{4}{c}{SD3~\cite{esser2024scaling}}
    & \multicolumn{4}{c}{FLUX.1-dev~\cite{labs2025flux}}
    & \multicolumn{4}{c}{Z-Image~\cite{cai2025z}} \\
    \cmidrule(lr){2-5}
    \cmidrule(lr){6-9}
    \cmidrule(lr){10-13}

    Method
    & Overall & Color & Shape & Texture
    & Overall & Color & Shape & Texture
    & Overall & Color & Shape & Texture \\
    \midrule
    
    Vanilla
    & 0.7111 & 0.8069 & 0.5986 & 0.7278
    & 0.6300 & 0.7609 & 0.5038 & 0.6253
    & 0.7292 & 0.8124 & 0.6141 & 0.7610 \\
    
    FK (flow-adapted)~\cite{singhal2025general}
    & 0.6793 & 0.7441 & 0.6036 & 0.6901
    & 0.5445 & 0.6283 & 0.4998 & 0.5054
    & 0.6455 & 0.6961 & 0.5926 & 0.6478 \\
    
    DATE~\cite{na2026diffusion}
    & 0.7153 & 0.8173 & 0.5986 & 0.7300
    & 0.6380 & 0.7686 & 0.5169 & 0.6284
    & 0.7310 & 0.8114 & 0.6155 & 0.7660 \\
    
    SMC~\cite{singhal2025general}
    & 0.7346 & \underline{0.8282} & 0.6218 & 0.7540
    & \underline{0.6723} & \underline{0.7884} & 0.5534 & \underline{0.6753}
    & 0.7431 & 0.8252 & \textbf{0.6438} & 0.7603 \\

    VP-SDE+RBF~\cite{kim2026inference}
    & 0.6902 & 0.7990 & 0.5690 & 0.7026
    & 0.6386 & 0.7633 & 0.5126 & 0.6398
    & 0.7201 & 0.8014 & 0.6136 & 0.7452 \\
    
    VP-SDE+SMC~\cite{kim2026inference}
    & \underline{0.7348} & 0.8274 & 0.6252 & 0.7517
    & 0.6631 & 0.7834 & 0.5458 & 0.6600
    & \textbf{0.7568} & \textbf{0.8348} & 0.6416 & \textbf{0.7939} \\
    
    NR-SMC~\cite{lian2026testtime}
    & 0.7337 & 0.8083 & \underline{0.6263} & \textbf{0.7664}
    & 0.6701 & 0.7847 & \underline{0.5546} & 0.6709
    & 0.7398 & 0.8145 & \underline{0.6360} & 0.7688 \\
    
    \midrule
    
    PromptSMC (Ours)
    & \textbf{0.7407} & \textbf{0.8381} & \textbf{0.6294} & \underline{0.7545}
    & \textbf{0.6887} & \textbf{0.8098} & \textbf{0.5707} & \textbf{0.6855}
    & \underline{0.7503} & \underline{0.8336} & 0.6342 & \underline{0.7829} \\

    \bottomrule
    \end{tabular}}
\end{table*}

\section{Additional Experiments}
\subsection{Template-based Prompt Rewriting}
\label{subsec:template_prompt}
Besides LLM-based rewriting, we also evaluate a lightweight template-based alternative that uses predefined text transformations designed to preserve the original meaning of the prompt. For example, given ``a photo of two red birds above a white dog'', the template-based rewriting may replace ``two'' with ``a pair of'' or ``above'' with ``directly above''. In contrast, the LLM-based rewriting can more freely restructure the sentence, \eg, ``A white dog is below two red birds'', while preserving the same semantic content.

\cref{app-tab:template} compares Vanilla, standard SMC, and our method using the two
rewriting approaches on SD3~\cite{esser2024scaling} and FLUX.1-dev~\cite{labs2025flux}. Notably, template-based rewriting outperforms both Vanilla and standard SMC across all metrics and base models, suggesting that even simple prompt transformations can improve prompt-image alignment. LLM-based rewriting further improves performance across all four metrics, showing the benefits of flexible prompt adaptation over fixed templates.

\subsection{Generalization Beyond SMC}
\label{subsec:dts}
We further apply prompt rejuvenation to Diffusion Tree Sampling (DTS)~\cite{jain2026diffusion}, a tree-based inference-time scaling method that explores different trajectories through stochastic denoising transitions. However, with deterministic flow-matching samplers, expanding the same node under a fixed prompt produces identical child states, limiting exploration. Our prompt rejuvenation approach addresses this issue by revising prompts across branches, allowing descendants of the same node to follow different conditional velocity fields. As shown in~\cref{tab:sd3_dts_rewriting}, at the same budget of 228 model function evaluations (NFE) per prompt on SD3, adding prompt rejuvenation to DTS improves both GenEval scores and pairwise LPIPS over vanilla DTS. These results show that prompt rejuvenation can improve both prompt-image alignment and within-prompt diversity beyond SMC.

\section{Additional Quantitative Results}
\label{sec:app_quantity}

\subsection{Task-wise Benchmark Results}
\label{subsec:task_wise}

We provide detailed task-wise results on GenEval and T2I-CompBench, complementing the overall results reported in the main paper (\cref{tab:geneval_main} and \cref{tab:t2i_attribute_main}). These results show how each method performs across different tasks and where the improvements come from.

\paragraph{GenEval.}

\cref{app-tab:geneval_breakdown_sd3},~\cref{app-tab:geneval_breakdown_flux} and~\cref{app-tab:geneval_breakdown_zimage}  report the task-wise performance on GenEval for SD3, FLUX.1-dev, and Z-Image, respectively. GenEval evaluates six tasks: single object, two objects, counting, colors, position, and color attribution. 
Our method achieves the best overall performance across all three backbones, with the best or tied-best results on most tasks. The largest gains are observed in two-object generation, position, and color attribution. Single-object generation is already close to perfect for most methods, leaving less room for improvement.

\paragraph{T2I-CompBench.}

We report attribute-binding performance on T2I-CompBench for SD3, FLUX.1-dev, and Z-Image in~\cref{tab:t2i_attribute_breakdown}. Specifically, we provide B-VQA scores for the three tasks (color, shape and texture) and their average as the overall score.
Our method achieves the best overall score on SD3 and FLUX, and the second-best overall score on Z-Image. On FLUX.1-dev, our method performs best on all three tasks. On SD3, it achieves the best color and shape scores, while remaining competitive on texture. On Z-Image, our method gives strong color and texture results, while VP-SDE+SMC performs slightly better overall.

\begin{table}[t]
\centering
\small
\renewcommand{\arraystretch}{1.1}
\caption{
Particle scaling results on GenEval using SD3. All quality results use ImageReward-based Top-1 selection. 
}
\vspace{-0.1in}
\label{tab:particle_scaling}

\resizebox{\columnwidth}{!}{%
\begin{tabular}{
@{}lcc>{\columncolor{genevalgreen}}ccc@{}
}
\toprule
Method
& $N$
& ImageReward$^{\dagger}$ $\uparrow$
& GenEval $\uparrow$
& Sec/prompt $\downarrow$
& Memory (GB) $\downarrow$ \\
\midrule

\multirow{5}{*}{Best-of-$N$}
& 2  & 1.2588 & 0.7531  & 7.17 & 28.19 \\
& 4  & 1.3703 & 0.7702  & 14.10 & 36.81 \\
& 8  & 1.4434 & 0.7914  & 27.80 & 36.91 \\
& 16 & 1.4827 & 0.7729  & 53.89 & 37.11 \\
& 32 & 1.5261 & 0.7836  & 107.36 & 37.51 \\

\midrule

\multirow{5}{*}{SMC}
& 2  & 1.2373 & 0.7442  & 9.56 & 28.19 \\
& 4  & 1.3458 & 0.7773  & 18.35 & 36.81  \\
& 8  & 1.4000 & 0.7804  & 35.39 & 36.91 \\
& 16 & 1.4488 & 0.7675 & 73.76	 & 37.11 \\
& 32 & 1.4953 & 0.7743  & 144.33 & 37.52 \\

\midrule

\multirow{5}{*}{PromptSMC}
& 2  & 1.2885 & 0.7988  & 10.67 & 45.72 \\
& 4  & 1.4068 & 0.8022  & 23.29 & 54.34 \\
& 8  & 1.4584 & 0.8079  & 46.94 & 54.45 \\
& 16 & 1.5032 & 0.8105  & 100.62 & 54.66 \\
& 32 & 1.5414 & 0.8033  & 210.22 & 55.06 \\

\bottomrule
\end{tabular}}
\end{table}

\subsection{Additional Results on Particle Scaling}
\label{subsec:app_particle}

We provide the full particle scaling results in~\cref{tab:particle_scaling}, complementing the scaling trends shown in \cref{fig:particle_scaling} of the main paper. We compare Best-of-$N$, standard SMC, and our method under identical experimental settings, with $N$ increasing from 2 to 32. All three methods use ImageReward-based top-1 selection. For each prompt, we select the final image with the highest ImageReward among the $N$ candidates and use the selected image to compute both reported quality metrics.

Our method consistently achieves the highest ImageReward and GenEval scores across all evaluated particle budgets. We also report wall-clock time and peak memory usage to show how computational cost changes with the number of particles.

\subsection{Resampling Settings and Guidance Strength}
\label{subsec:app_resample}

We present detailed results on the effects of resampling window, interval and guidance strength in~\cref{app-tab:rewrite_window},~\cref{app-tab:rewrite_frequency} and ~\cref{app-tab:lambda_ablation}, respectively.
The shaded rows indicate the default settings used in our main experiments.
ImageReward$^\dagger$ is the reward  function used for particle guidance,
and higher LPIPS indicates greater within-prompt diversity.

\paragraph{Resampling window.}

\cref{app-tab:rewrite_window} studies when resampling is applied during generation. We use 50 sampling steps and vary the position of a 30-step resampling window, while keeping all other settings fixed. We also consider the full 50-step window for comparison.
We observe that resampling at later stages leads to higher LPIPS, indicating greater within-prompt diversity, but lower GenEval scores. In contrast, earlier windows reduce diversity without further improving GenEval over the default window [10, 40). Notably, resampling throughout the entire sampling process does not provide additional benefits.

\begin{table}[htbp]
\centering
\small
\caption{
Effect of the prompt-rewriting window on GenEval using SD3. The shaded row denotes our default setting. }
\vspace{-0.1in}
\label{app-tab:rewrite_window}
\resizebox{\columnwidth}{!}{
\begin{tabular}{@{}lccccc@{}}
\toprule
Window
& ImageReward $\uparrow$
& GenEval $\uparrow$
& B-CLIP $\uparrow$
& VQAScore $\uparrow$
& LPIPS $\uparrow$ \\
\midrule

$[0,30)$
& 1.3152
& 0.7842
& 0.8012
& 0.9008
& 0.1752 \\

$[5,35)$
& 1.3180
& 0.7824
& 0.8015
& \textbf{0.9065}
& 0.1681 \\

\rowcolor{defaultrow}
$[10,40)$
& \textbf{1.3331}
& \textbf{0.7899}
& 0.8026
& 0.9038
& 0.2437 \\

$[15,45)$
& 1.3188
& 0.7720
& \textbf{0.8029}
& 0.8983
& 0.3221 \\

$[20,50)$
& 1.3244
& 0.7743
& 0.8008
& 0.9059
& \textbf{0.3403} \\

$[0,50)$
& 1.3205
& 0.7819
& 0.8023
& 0.8970
& 0.1769 \\

\bottomrule
\end{tabular}
}
\end{table}

\paragraph{Resampling interval.}

We also examine how resampling interval affects generation quality and diversity in~\cref{app-tab:rewrite_frequency}. Notably, more frequent resampling is more costly but does not improve the generation quality. Moreover, it also reduces the diversity of the generated images. Less frequent resampling reduces runtime and increases diversity, but slightly lowers GenEval scores compared with the default setting ($f=5$).

\begin{table}[htbp]
\centering
\caption{
Effect of the prompt-rewriting interval $f$ on GenEval using SD3. The shaded row denotes our default setting.
}
\vspace{-0.1in}
\label{app-tab:rewrite_frequency}

\resizebox{\columnwidth}{!}{
\begin{tabular}{@{}cccccc@{}}
\toprule
Interval
& ImageReward $\uparrow$
& GenEval $\uparrow$
& LPIPS $\uparrow$
& Time (s/prompt) $\downarrow$ \\
\midrule

1
& 1.3132
& 0.7713
& 0.2282
& 37.53 \\

\rowcolor{defaultrow}
5
& \textbf{1.3331}
& \textbf{0.7899}
& 0.2437
& 23.29 \\

10
& 1.3195
& 0.7836
& 0.2506
& 21.70\\

15
& 1.3120
& 0.7844
& \textbf{0.2671}
& \textbf{21.61}\\

\bottomrule
\end{tabular}
}
\end{table}

\paragraph{Guidance strength.}

We study the effect of guidance strength $\lambda$ in~\cref{app-tab:lambda_ablation}. Increasing $\lambda$ initially improves GenEval and ImageReward, but also progressively reduces LPIPS, indicating a trade-off between reward-guided selection and within-prompt diversity. We use $\lambda=10$ as the default setting, which provides a balance between generation quality and diversity.

\begin{table}[htbp]
\centering
\caption{
Effect of the guidance strength $\lambda$ on GenEval using SD3. The shaded row denotes our default setting. 
}
\vspace{-0.1in}
\label{app-tab:lambda_ablation}

\resizebox{\columnwidth}{!}{
\begin{tabular}{@{}cccccc@{}}
\toprule
$\lambda$
& ImageReward $\uparrow$
& GenEval $\uparrow$
& B-CLIP $\uparrow$
& VQAScore $\uparrow$
& LPIPS $\uparrow$ \\
\midrule

0
& 1.0912
& 0.7375
& 0.7907
& 0.8953
& \textbf{0.6665} \\

5
& 1.2562
& 0.7572
& 0.8000
& 0.9039
& 0.4068 \\

\rowcolor{defaultrow}
10
& 1.3331
& 0.7899
& \textbf{0.8026}
& 0.9038
& 0.2437 \\

20
& 1.3405
& \textbf{0.8004}
& 0.8008
& 0.9036
& 0.1341 \\

30
& \textbf{1.3478}
& 0.7933
& 0.8007
& \textbf{0.9078}
& 0.0926 \\

\bottomrule
\end{tabular}
}
\end{table}

\subsection{Additional Performance-Efficiency Analysis}
\label{subsec:efficient}
We compare PromptSMC with DATE~\cite{na2026diffusion}, as both methods adapt text conditioning during sampling. DATE updates prompt embeddings through reward gradients, whereas PromptSMC rewrites the prompt text using LLMs. Like the other baselines, DATE generates four candidates per prompt. In \cref{tab:geneval_main,tab:t2i_attribute_main}, we use DATE’s full-update setting (DATE 100\%), which updates the prompt embeddings at every sampling step. In~\cref{tab:date_compute}, we also report results for DATE with updates at 10\%, 20\%, and 50\% of the 50 sampling steps. PromptSMC has a similar compute cost to DATE (10\%) but achieves substantially higher GenEval scores. DATE (100\%) requires 100 more NFE and takes 4.6 times as long as PromptSMC, but its GenEval score is still much lower than PromptSMC. 

\begin{table}[ht]
\centering

\caption{
Compute–performance comparison of DATE and PromptSMC on GenEval with SD3. NFE, runtime, and peak GPU memory are reported per prompt.
}
\label{tab:date_compute}

\resizebox{\columnwidth}{!}{
\begin{tabular}{@{}lc>{\columncolor{genevalgreen}}cccccc@{}}
\toprule
Method
& ImageReward
& GenEval
& B-CLIP
& VQAScore
& NFE
& Time (s)
& Memory (GB) \\
\midrule

Vanilla
& 1.0674
& 0.7214
& 0.7885
& 0.8886
& 200
& 14.10
& 35.01 \\

DATE (10\%)
& 1.0676
& 0.7243
& 0.7871
& 0.8878
& 210
& 21.55
& 44.98 \\

DATE (20\%)
& 1.0794
& 0.7189
& 0.7881
& 0.8912
& 220
& 30.60
& 44.98 \\

DATE (50\%)
& 1.1000
& 0.7191
& 0.7879
& 0.8888
& 250
& 58.31
& 44.98 \\

DATE (100\%)
& 1.1397
& 0.7365
& 0.7872
& 0.8868
& 300
& 106.72
& 44.98 \\

\midrule

PromptSMC (Ours)
& \textbf{1.3331}
& \textbf{0.7899}
& \textbf{0.8026}
& \textbf{0.9038}
& 200
& 23.29
& 54.34 \\

\bottomrule
\end{tabular}
}
\end{table}

\begin{figure*}[ht]
  \centering
  \includegraphics[width=1.\textwidth]{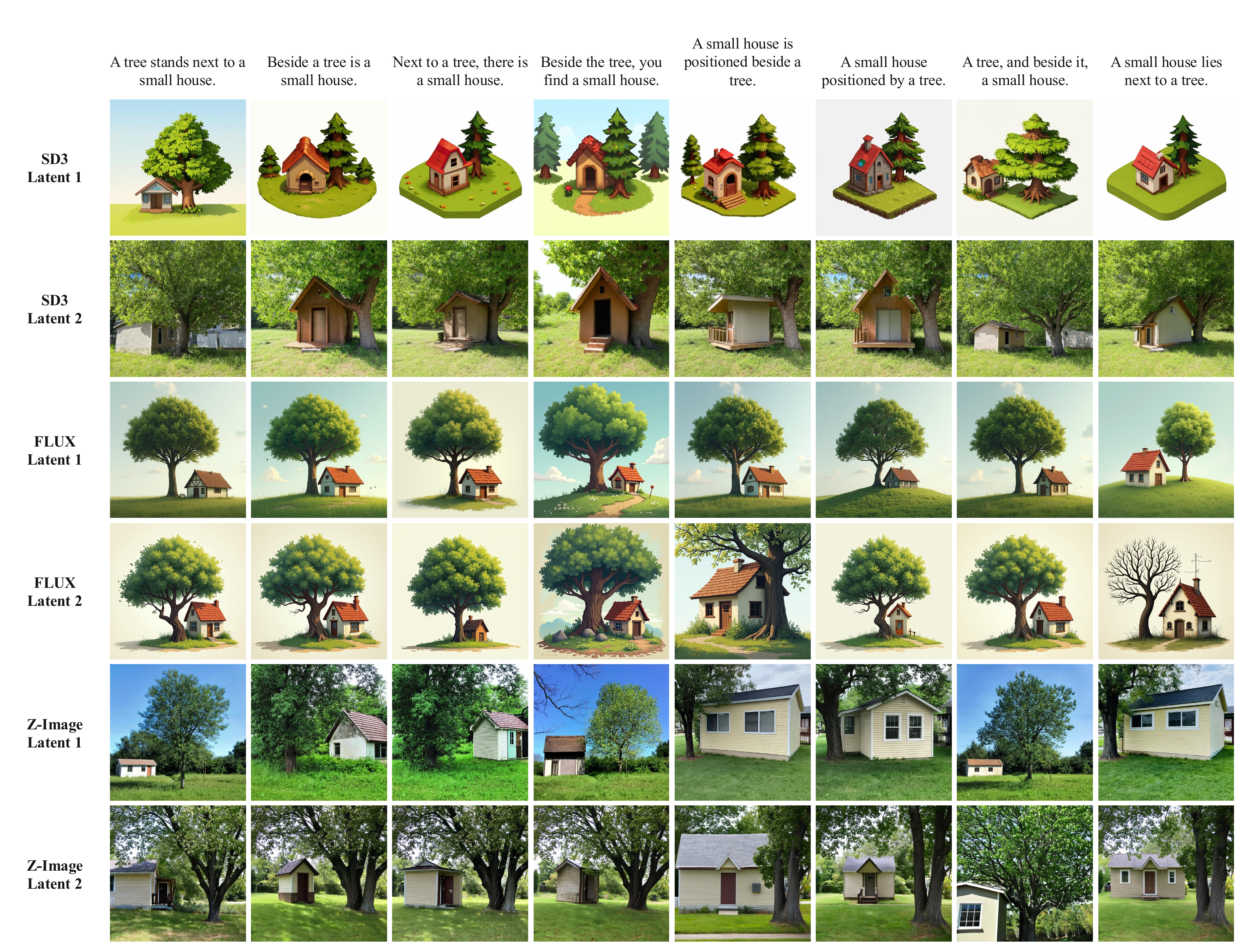}
    \caption{\textbf{Effect of prompt paraphrasing on image generation.} Each row corresponds to a fixed backbone and initial latent. Each column represents a different paraphrase of the same prompt.  Within each row, the initial latent and sampling settings are fixed, but different prompt paraphrases can still produce noticeably different images.}
    \label{app-fig:same_latent}
    \vspace{-0.1in}
\end{figure*}

\section{Additional Qualitative Results}
\label{sec:more_quality}
\subsection{Prompt Rewriting with a Fixed Latent}

To motivate prompt-space exploration, we examine whether paraphrasing the prompt alone can alter the generation outcome when the initial latent is fixed.
As shown in Fig.~\ref{app-fig:same_latent}, even with the initial latent and sampling settings fixed, semantically similar prompt rephrasings can still lead to different images, consistently across three backbones. This observation suggests that prompt rewriting can provide a simple way to induce alternative generation paths without perturbing the latent state.

\subsection{Comparison with Baselines}
\label{sec:quality}
\begin{figure*}[t]
    \centering
    \includegraphics[width=0.95\linewidth]{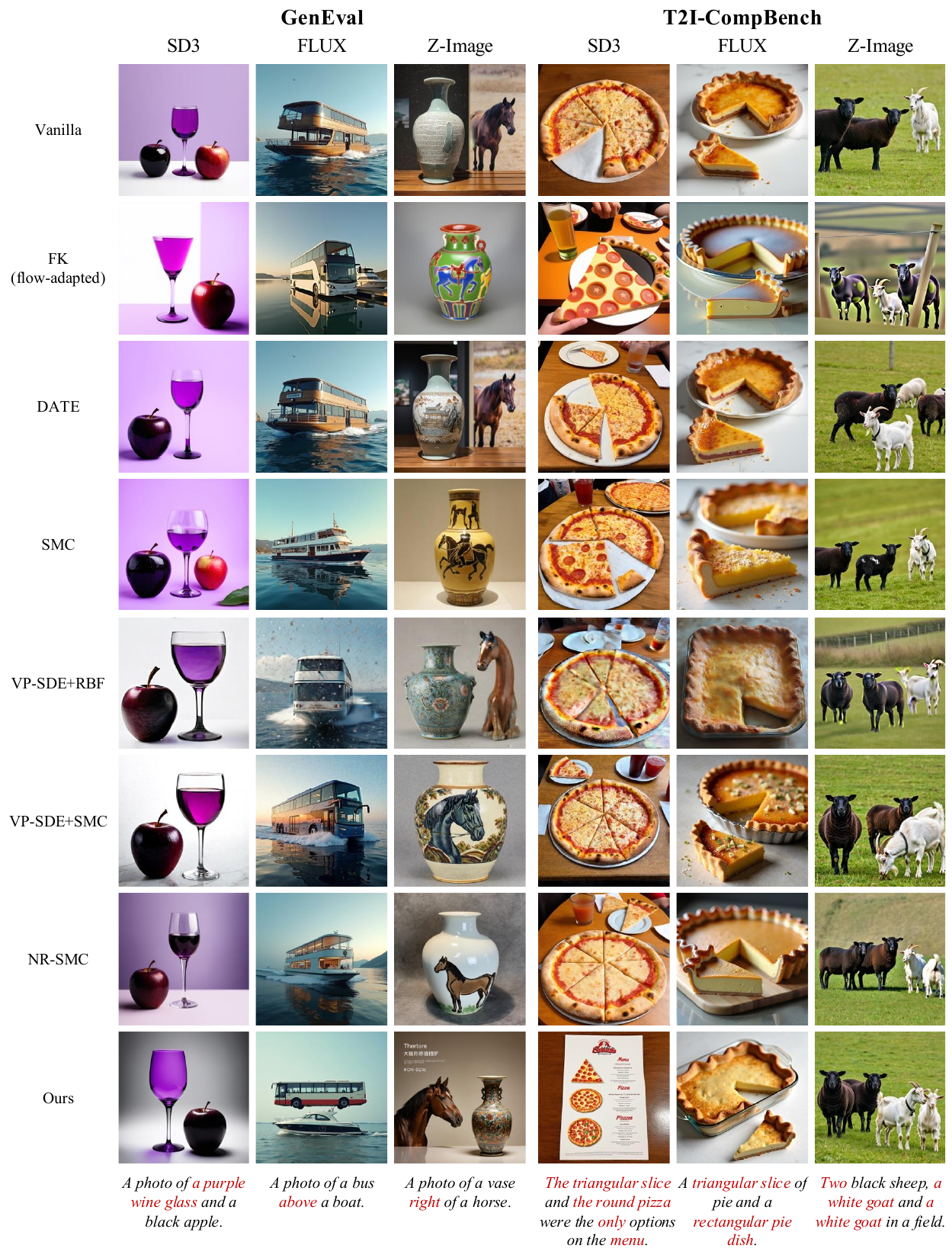}
    \caption{Visual comparison of our method with other state-of-the-art inference-time scaling approaches on GenEval and T2I-CompBench. Key requirements in each prompt are highlighted in red. Please zoom in to view the details.}
    \label{app-fig:more_both}
\end{figure*}

\begin{figure*}[t]
    \centering
    \includegraphics[width=0.65
    \linewidth]{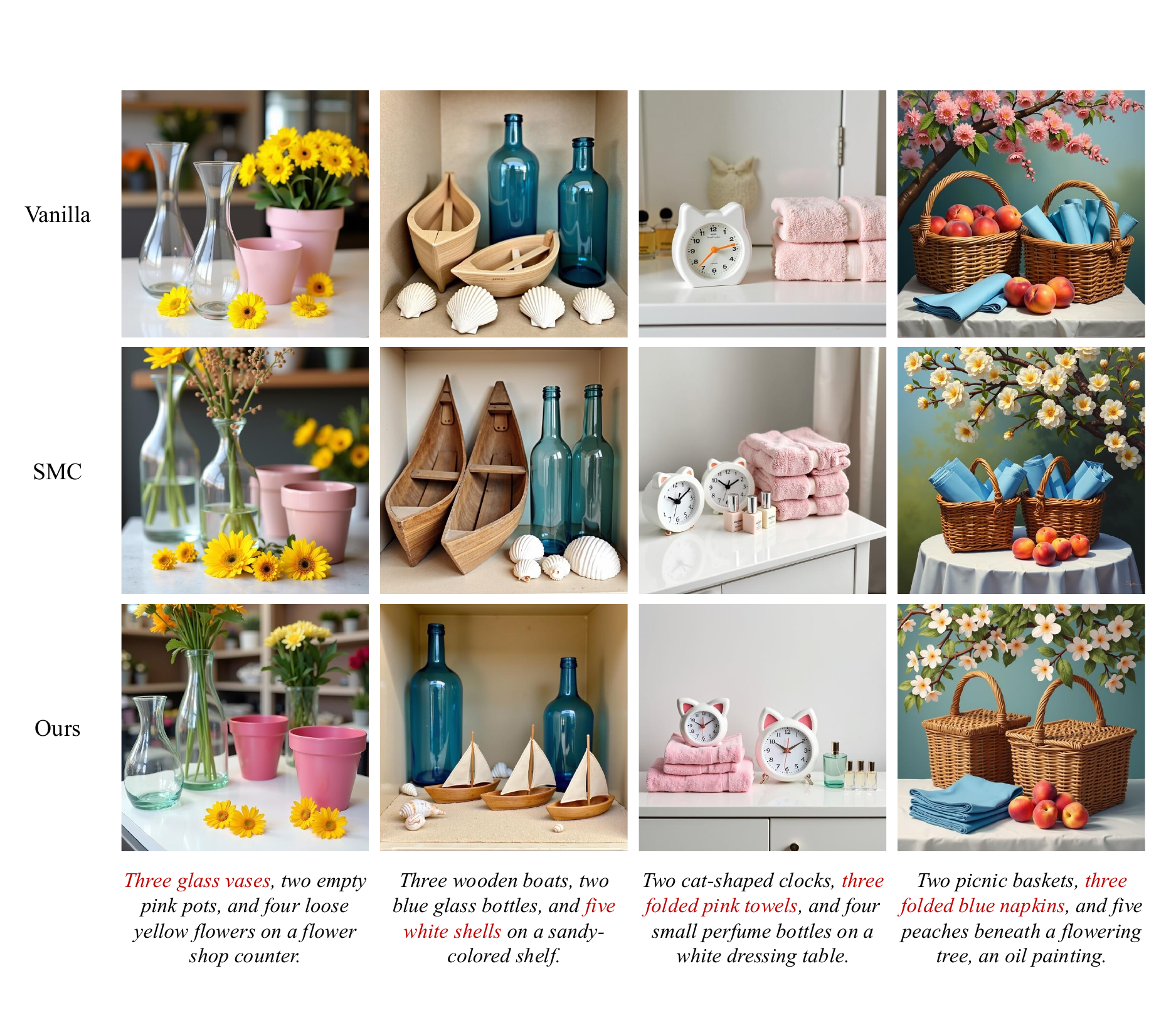}
    \caption{Qualitative comparison of vanilla generation, standard SMC, and PromptSMC on prompts involving complex counting using the FLUX.1-dev backbone. Key prompt requirements are highlighted in red. Zoom in for details.}
    \label{app-fig:more4}
\end{figure*}

\begin{figure*}[t]
    \centering
    \includegraphics[width=0.65\linewidth]{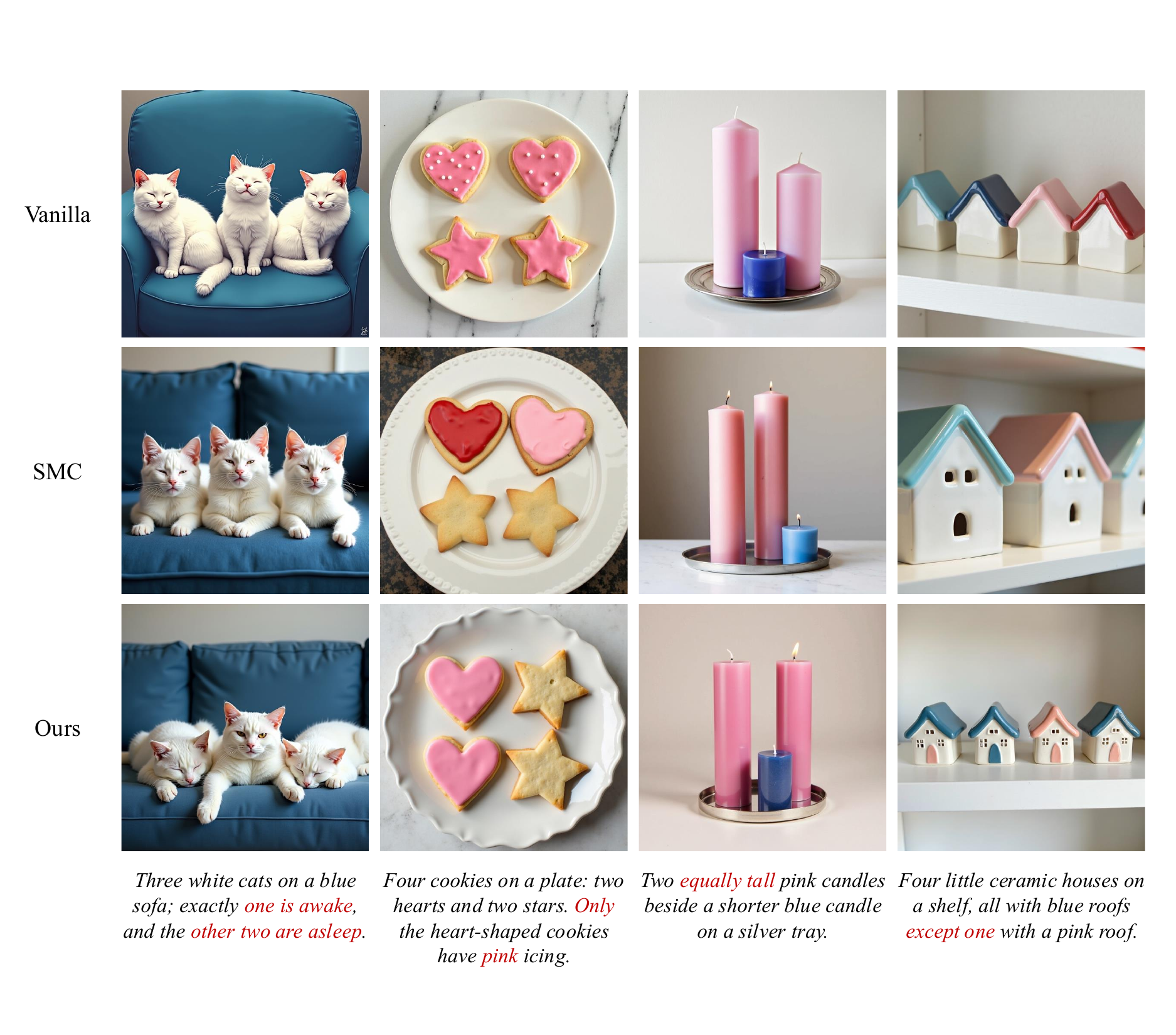}
    \caption{Qualitative comparison of vanilla generation, standard SMC, and PromptSMC on prompts involving logical constraints using the FLUX.1-dev backbone. Key prompt requirements are highlighted in red. Zoom in for details.}
    \label{app-fig:more3}
\end{figure*}

\begin{figure*}[t]
    \centering
    \includegraphics[width=0.65\linewidth]{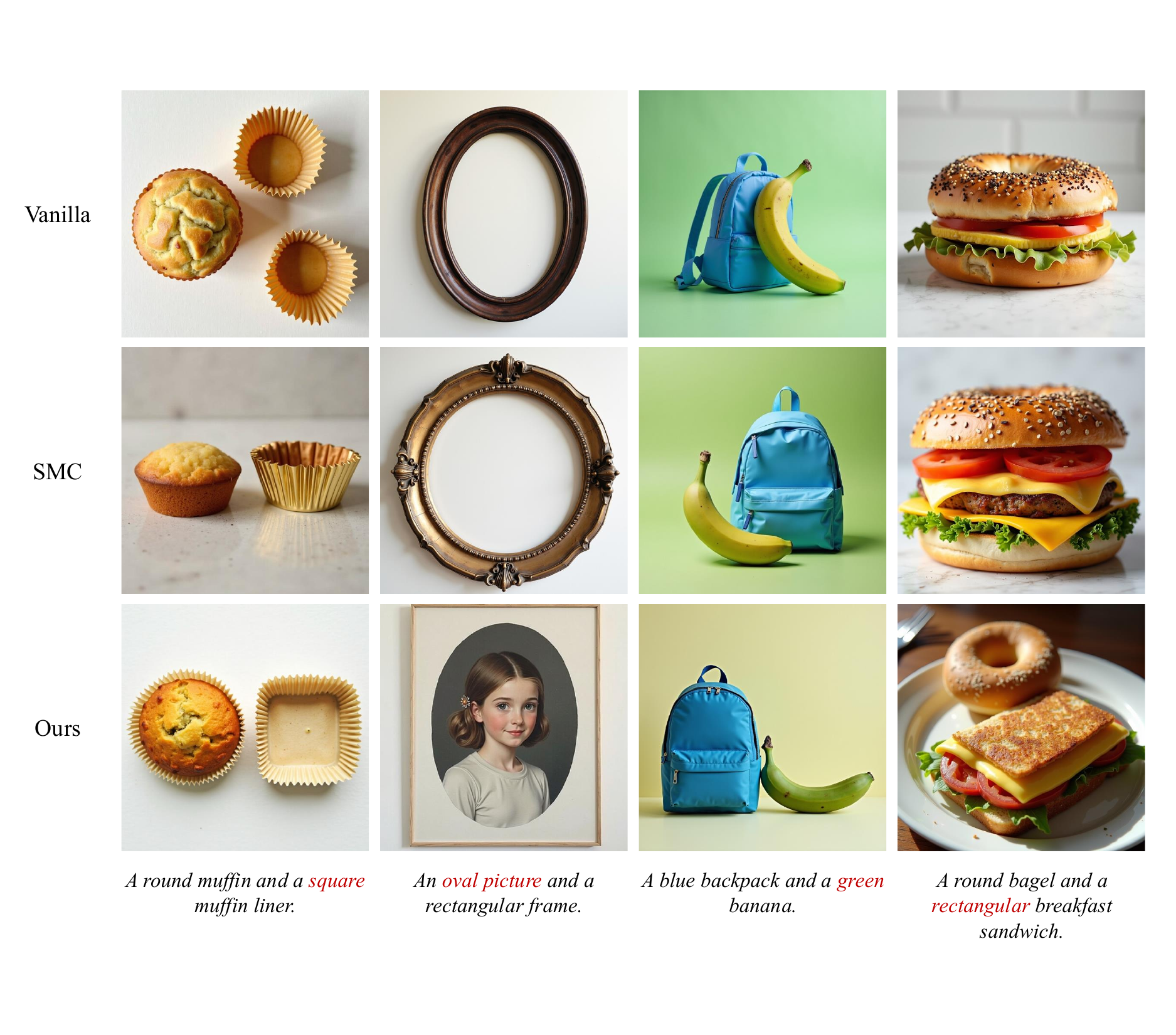}
    \caption{Qualitative comparison of vanilla generation, standard SMC, and PromptSMC on challenging attribute-binding prompts using the FLUX.1-dev backbone. Key prompt requirements are highlighted in red. Zoom in for details.}
    \label{app-fig:more1}
\end{figure*}

\begin{figure*}[t]
    \centering
    \includegraphics[width=0.65\linewidth]{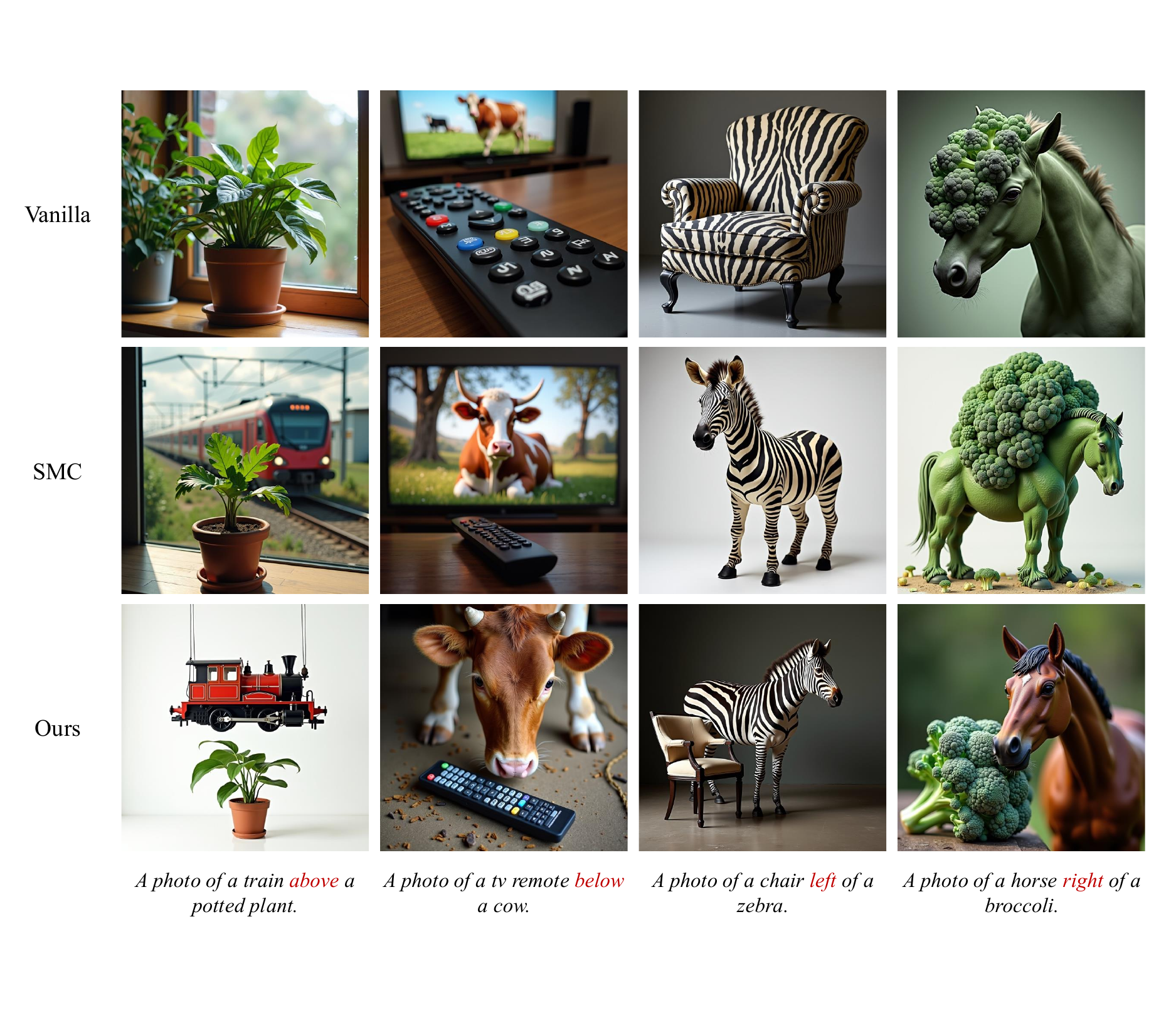}
    \caption{Qualitative comparison of vanilla generation, standard SMC, and PromptSMC on prompts involving unusual spatial relationships using the FLUX.1-dev backbone. Key prompt requirements are highlighted in red. Zoom in for details.}
    \label{app-fig:more2}
\end{figure*}

We provide additional qualitative comparisons with other state-of-the-art inference-time scaling methods on both benchmarks in~\cref{app-fig:more_both}.
We also present additional examples with more challenging prompts in \cref{app-fig:more4,app-fig:more3,app-fig:more1,app-fig:more2}. 
These figures compare vanilla generation, standard SMC, and PromptSMC using FLUX.1-dev. They illustrate complex counting, logical constraints, challenging attribute binding, and unusual spatial relationships, respectively.

\subsection{Failure Cases and Limitations}
\label{subsec:failure_cases}
\begin{figure*}[t]
    \centering
    \includegraphics[width=0.67\linewidth]{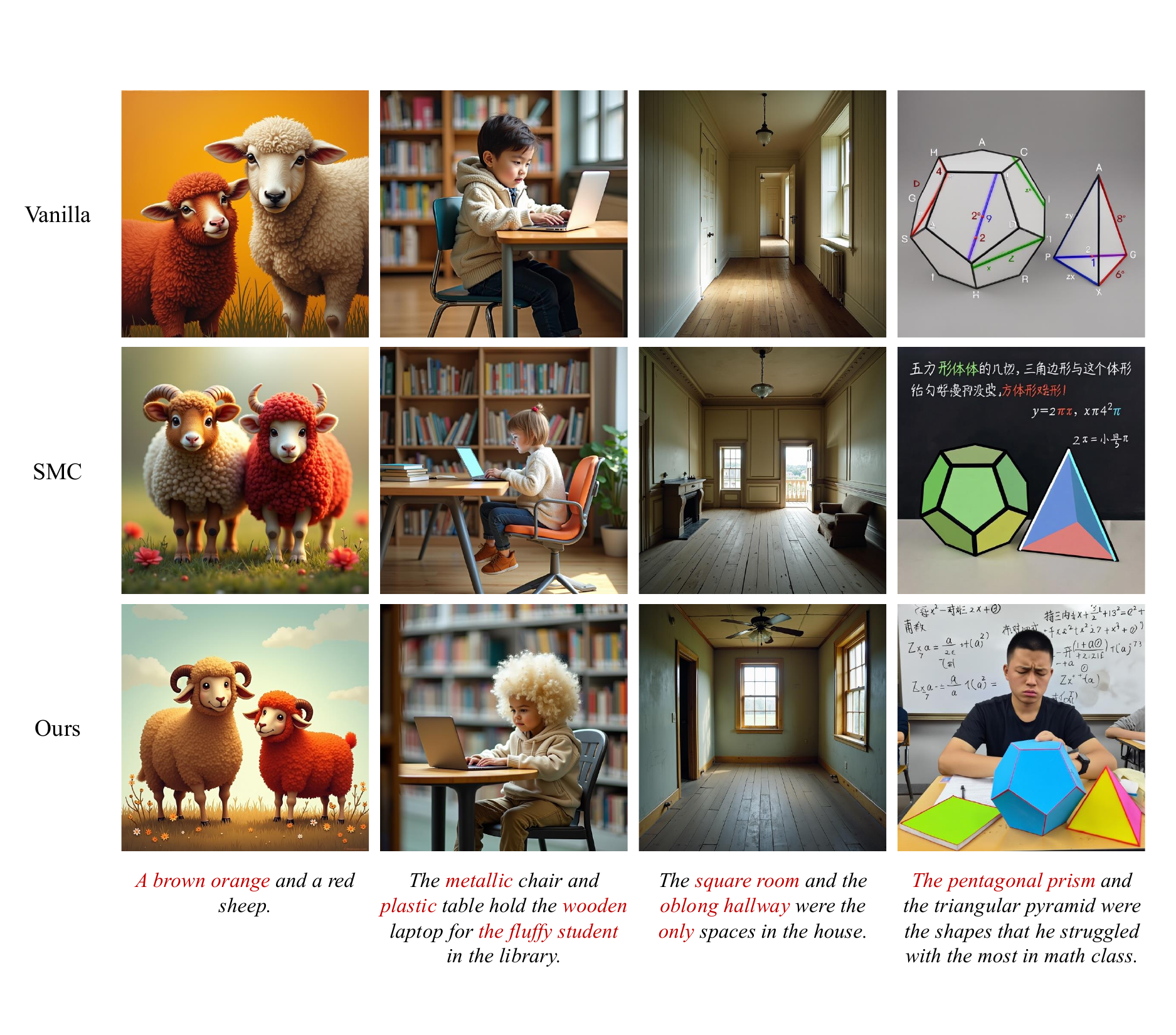}
    \caption{Failure cases of our method involving lexical ambiguity, unrealistic attribute combinations, scene-level constraints and complex geometry. Key prompt requirements are highlighted in red.}
    \label{app-fig:fail}
\end{figure*}

Here, we analyze different types of failure cases of our method. On both benchmarks, we observe failures caused by lexical ambiguity, where models misinterpret words with multiple meanings. For example, some prompts contain the word ``orange", which can refer to either a color or a fruit. Given the prompt ``A brown orange and a red sheep", both the baselines and our method typically generate a brown sheep and a red sheep, rather than a brown orange and a red sheep, as shown in the first column of~\cref{app-fig:fail}.
Some prompts contain constraints that are difficult to verify from a single image. For instance, given a prompt ``The square room and the oblong hallway were the only spaces in the house", it is difficult to determine whether the generated images depict the correct shapes due to perspective and occlusion, as shown in the third column of~\cref{app-fig:fail}. Moreover, it is difficult to verify whether these are the \emph{only} spaces in the house from a single image. Therefore, evaluating such prompts can be challenging and unreliable.
In addition, prompt rejuvenation relies on pretrained LLMs, and the rewritten prompts may occasionally deviate from the original semantics, potentially affecting generation quality. The effectiveness of our method may also depend on how well the reward model captures the desired prompt constraints. Finally, LLM-based prompt rewriting introduces additional runtime and memory overhead, which may not be justified for relatively simple prompts. 

\end{document}